\documentclass[dvipsnames]{article} %
\usepackage{colm2024_conference}

\usepackage{booktabs}
\usepackage{graphicx}
\usepackage{enumitem}
\usepackage{wrapfig}
\usepackage{algorithm}
\usepackage{algpseudocode}
\usepackage{amsmath}
\usepackage{colortbl}
\usepackage[utf8]{inputenc}
\usepackage{caption}
\usepackage{multirow}
\usepackage{tabularx}
\usepackage{pgfplots}
\pgfplotsset{compat=1.18}
\usepackage{fontawesome5}
\usetikzlibrary{positioning,calc,arrows.meta,shapes.geometric,fit,backgrounds}
\usepackage{makecell}
\usepackage{siunitx}
\usepackage{nicefrac}
\usepackage{tocloft}
\usepackage{listings}
\usepackage[raster,skins]{tcolorbox}
\usepackage{xltabular}
\usepackage{adjustbox}
\usepackage{xurl}
\usepackage{multicol}
\usepackage{rotating}
\usepackage{cleveref}
\crefname{section}{\S}{\S\S}
\Crefname{section}{\S}{\S\S}
\usepackage[normalem]{ulem}
\useunder{\uline}{\ul}{}
\usepackage{tablefootnote}

\usepackage{amsmath,amsfonts,bm}

\def\eqref#1{equation~\ref{#1}}

\def\1{\bm{1}}

\DeclareMathAlphabet{\mathsfit}{\encodingdefault}{\sfdefault}{m}{sl}
\SetMathAlphabet{\mathsfit}{bold}{\encodingdefault}{\sfdefault}{bx}{n}

\usepackage{float}
\usepackage{amsfonts}
\usepackage{amssymb}
\usepackage{array}
\usepackage{comment}
\usepackage{hhline}
\usepackage{boldline}
\usepackage{subfig}
\usepackage{framed}
\usepackage{color}
\usepackage{pifont}
\tcbuselibrary{skins, breakable, listings}

\definecolor{posgreen}{HTML}{E8F5E9}
\definecolor{negred}{HTML}{FFEBEE}
\definecolor{neublue}{HTML}{E3F2FD}

\newcommand\eat[1]{}

\newcommand{\cmark}{\ding{51}}
\newcommand{\xmark}{\ding{55}}
\usepackage{iftex}
\ifPDFTeX
  \usepackage{CJKutf8}
  
\fi
\newcommand{\score}[2]{\ensuremath{#1_{\scriptscriptstyle #2}}}
\newcommand{\bestscore}[2]{\textcolor{promptcardgreendark}{\ensuremath{\mathbf{#1}_{\scriptscriptstyle #2}}}}
\newtcolorbox{mechheader}{colback=black!4,colframe=black!60,boxrule=0.4pt,arc=1.5pt,
  left=5pt,right=5pt,top=3pt,bottom=3pt,before skip=10pt,after skip=6pt}
\definecolor{promptaccent}{HTML}{315D6E}
\definecolor{promptpaper}{HTML}{F7F9F9}
\definecolor{promptrule}{HTML}{C6CED1}
\definecolor{promptcardblue}{HTML}{2A78D6}
\definecolor{promptcardbluedark}{HTML}{1C5CAB}
\definecolor{promptcardnavy}{HTML}{0D366B}
\definecolor{promptcardgreen}{HTML}{1BAF7A}
\definecolor{promptcardgreendark}{HTML}{12805A}
\definecolor{promptcardbluebg}{HTML}{EAF1FC}
\definecolor{promptcardgraybg}{HTML}{F3F2EE}
\definecolor{promptcardgreenbg}{HTML}{E7F6F0}
\definecolor{promptcardgray}{HTML}{898781}
\definecolor{promptcardsurface}{HTML}{FCFCFB}
\definecolor{promptcardline}{HTML}{E1E0D9}
\definecolor{q4revisionborder}{HTML}{C84C4C}
\definecolor{q4conflictborder}{HTML}{16845E}
\definecolor{tblzebra}{HTML}{F8F8F5}
\definecolor{tblbandgray}{HTML}{EFEEE9}
\newcommand{\tbest}[1]{\textcolor{promptcardgreendark}{\textbf{#1}}}
\newcommand{\tsecond}[1]{\textcolor{promptcardbluedark}{\underline{#1}}}
\newcommand{\tblock}[3]{\multicolumn{#1}{@{}l}{\textit{\textcolor{#2}{\textbf{#3}}}}}
\newcommand{\tyes}{\textcolor{promptcardgreendark}{\cmark}}
\newcommand{\tno}{\textcolor{promptcardgray}{\xmark}}
\newtcblisting{promptbox}[1]{enhanced,breakable,listing only,listing engine=listings,
  title={\textcolor{promptaccent}{\rmfamily\scshape\normalsize Prompt}%
    \hspace{0.8em}{\rmfamily\mdseries\normalsize #1}},
  fonttitle=\normalfont,coltitle=black,colbacktitle=white,
  colback=promptpaper,colframe=promptrule,boxrule=0.35pt,arc=1pt,
  borderline west={1.2pt}{0pt}{promptaccent},titlerule=0.35pt,
  left=6pt,right=6pt,top=4pt,bottom=4pt,
  lefttitle=6pt,righttitle=6pt,toptitle=3pt,bottomtitle=3pt,
  before skip=6pt,after skip=8pt,
  listing options={basicstyle=\ttfamily\normalsize,breaklines=true,
    breakautoindent=false,breakindent=0pt,columns=fullflexible,keepspaces=true,
    showstringspaces=false}}
\newtcolorbox{structuredprompt}[1]{enhanced,breakable,
  colframe=promptcardline,colback=promptcardsurface,boxrule=0.5pt,arc=1.5pt,
  colbacktitle=promptcardnavy,coltitle=white,
  left=2pt,right=2pt,top=12pt,bottom=1pt,
  before skip=3pt,after skip=4pt,
  attach boxed title to top left={xshift=5pt,yshift=-\tcboxedtitleheight/2},
  boxed title style={colframe=promptcardnavy,colback=promptcardnavy,
    boxrule=0pt,arc=1.5pt,left=5pt,right=5pt,top=2pt,bottom=2pt},
  title={\textcolor{white}{\sffamily\bfseries\small PROMPT\enspace
    \normalfont #1}},
  title after break={\textcolor{white}{\sffamily\bfseries\small PROMPT\enspace
    \normalfont #1\enspace\textit{(continued)}}}}
\newtcolorbox{promptregion}[4]{enhanced,breakable=false,
  colback=#1,boxrule=0pt,arc=1.5pt,
  borderline west={1.8pt}{0pt}{#2},
  left=5pt,right=4pt,top=10.5pt,bottom=1pt,
  before skip=1.5pt,after skip=1.5pt,
  overlay={\node[anchor=north west,inner sep=0pt,
    font=\sffamily\bfseries\scriptsize,text=#3]
    at ([xshift=5pt,yshift=-3pt]frame.north west){#4};}}
\newtcblisting{promptlistingregion}[4]{enhanced,breakable=false,
  listing only,listing engine=listings,
  colback=#1,boxrule=0pt,arc=1.5pt,
  borderline west={1.8pt}{0pt}{#2},
  left=5pt,right=4pt,top=10.5pt,bottom=1pt,
  before skip=1.5pt,after skip=1.5pt,
  overlay={\node[anchor=north west,inner sep=0pt,
    font=\sffamily\bfseries\scriptsize,text=#3]
    at ([xshift=5pt,yshift=-3pt]frame.north west){#4};},
  listing options={basicstyle=\ttfamily\footnotesize,breaklines=true,
    breakatwhitespace=true,breakautoindent=false,breakindent=0pt,
    aboveskip=0pt,belowskip=0pt,prebreak={},postbreak={},
    columns=fullflexible,keepspaces=true,
    showstringspaces=false}}
\newcommand{\promptbody}{\ttfamily\footnotesize\raggedright
  \setlength{\baselineskip}{10.5pt}}
\newtcblisting{artifactbox}[1]{enhanced,breakable,listing only,listing engine=listings,
  title={\rmfamily\mdseries\normalsize #1},
  fonttitle=\normalfont,coltitle=black,colbacktitle=white,
  colback=promptpaper,colframe=promptrule,boxrule=0.35pt,arc=1pt,
  borderline west={1.2pt}{0pt}{promptaccent},titlerule=0.35pt,
  left=6pt,right=6pt,top=4pt,bottom=4pt,
  lefttitle=6pt,righttitle=6pt,toptitle=3pt,bottomtitle=3pt,
  before skip=6pt,after skip=8pt,
  listing options={basicstyle=\ttfamily\normalsize,breaklines=true,
    breakautoindent=false,breakindent=0pt,columns=fullflexible,keepspaces=true,
    showstringspaces=false}}
\definecolor{treefolder}{HTML}{C58B26}
\definecolor{treetypescript}{HTML}{3178C6}
\definecolor{treejson}{HTML}{B58A17}
\definecolor{treejavascript}{HTML}{B88A16}
\definecolor{treetest}{HTML}{3F8F65}
\definecolor{treefile}{HTML}{76828B}
\definecolor{treetext}{HTML}{27323A}
\newcommand{\treeicon}[2]{%
  \makebox[9pt][c]{\textcolor{#1}{\scriptsize #2}}\hspace{2pt}}
\newcommand{\treefoldernode}[1]{%
  \treeicon{treefolder}{\faFolder}\textcolor{treetext}{\ttfamily #1}}
\newcommand{\treetsnode}[1]{%
  \treeicon{treetypescript}{\faFileCode}\textcolor{treetext}{\ttfamily #1}}
\newcommand{\treejsonnode}[1]{%
  \treeicon{treejson}{\faFileCode}\textcolor{treetext}{\ttfamily #1}}

\newcommand{\treetestnode}[1]{%
  \treeicon{treetest}{\faVial}\textcolor{treetext}{\ttfamily #1}}
\newcommand{\treeconfignode}[1]{%
  \treeicon{treefile}{\faCog}\textcolor{treetext}{\ttfamily #1}}
\newcommand{\treegenericnode}[1]{%
  \treeicon{treefile}{\faFile}\textcolor{treetext}{\ttfamily #1}}
\newtcblisting{treebox}[3]{enhanced,breakable,listing only,listing engine=listings,
  title={\parbox{\linewidth}{%
    \begingroup\setlength{\fboxsep}{3pt}%
    \colorbox{promptcardnavy}{\textcolor{white}{\sffamily\bfseries\scriptsize #1}}%
    \hspace{5pt}{\sffamily\bfseries\small #2}\par
    \vspace{1.5pt}\textcolor{treefolder}{\scriptsize\faFolderOpen}\hspace{3pt}%
    \textcolor{promptcardgray}{\footnotesize\path{#3}}%
    \endgroup}},
  fonttitle=\normalfont,coltitle=black,colbacktitle=promptcardgraybg,
  colback=promptcardsurface,colframe=promptcardline,boxrule=0.5pt,arc=2pt,
  borderline west={1.8pt}{0pt}{promptcardblue},titlerule=0.35pt,
  left=6pt,right=5pt,top=3pt,bottom=3pt,
  lefttitle=6pt,righttitle=5pt,toptitle=4pt,bottomtitle=3pt,
  before skip=5pt,after skip=7pt,
  listing options={basicstyle=\ttfamily\footnotesize,breaklines=true,
    breakatwhitespace=true,breakautoindent=false,breakindent=0pt,
    aboveskip=0pt,belowskip=0pt,columns=fullflexible,keepspaces=true,
    showstringspaces=false,escapeinside={(*@}{@*)},
    literate={|--}{{\textcolor{promptcardgray}{|--}}}3
             {`--}{{\textcolor{promptcardgray}{`--}}}3
             {|}{{\textcolor{promptcardgray}{|}}}1}}

\title{Terminal-Universe: Turning Agent Trajectories\\ into Scalable Terminal Environments}

\author{Jie Wu\textsuperscript{1,2,*}, Zhenru Zhang\textsuperscript{1,\ddag}, Beichen Zhang\textsuperscript{1}, Xuwu Wang\textsuperscript{1}, Yuhui Su\textsuperscript{1}, \\ Mouxiang Chen\textsuperscript{1}, Peng Wang\textsuperscript{1}, Zhihai Wang\textsuperscript{1}, Que Shen\textsuperscript{1}, Hao Zhou\textsuperscript{1}, \\
An Yang\textsuperscript{1}, Fei Huang\textsuperscript{1}, Yujiu Yang\textsuperscript{2,\dag}, Dayiheng Liu\textsuperscript{1,\dag} \\[0.7em]
{\bf \textsuperscript{1}Qwen Team, Alibaba Group \qquad \textsuperscript{2}Tsinghua University} \\[0.35em]
\textsuperscript{\ddag}Project leader \qquad \textsuperscript{\dag}Corresponding authors
}

\begin{document}

\maketitle

{\renewcommand{\thefootnote}{\fnsymbol{footnote}}%
 \footnotetext[1]{Work done during the internship at Qwen Team, Alibaba Group.}}

\begin{abstract}
\begin{sloppypar}
As terminal-based code agents become increasingly prevalent, the corresponding agent trajectories have accumulated at scale, while realistic, executable environments remain scarce.
However, environments are what agent post-training actually requires: each one can be re-queried into many verifiable tasks and provides the execution feedback, whereas a trajectory is a single frozen demonstration. 
Rather than generating environments from scratch, we observe that the tool-execution history in the existing agent trajectories exposes the structure and contents of the environments in which they ran, making it possible to reconstruct those environments from the trajectories themselves.
Thus, we introduce \textbf{Terminal-Universe}, a framework which turns each trajectory into a reusable environment and explores it for synthesizing new tasks and continued interactions. 
Specifically, Terminal-Universe replays the file operations recorded in a trajectory to restore each file before the agent modified it, yielding a partial workspace; a completion agent then supplies the missing files and dependencies. 
On this recovered workspace, we both reconstruct the original intent task and synthesize entirely new ones.
Besides, we also scale the tasks for the corresponding environment along two complementary axes: breadth and depth, to reproduce a routine pattern of real engineering practice. 
For breadth, we mine directional dependency relations between related environments and synthesize cross-workspace queries spanning multiple codebases, as developers routinely do in real-world development.
For depth, we extend the initial single-turn query into a multi-round session that captures iterative user feedback and requirement refinement via a user agent.
Applied to publicly available terminal agent trajectories, Terminal-Universe produces $37.3$k task-sufficient environments. 
Supervised fine-tuning of Qwen3.5-27B on this corpus improves single-round performance on Terminal-Bench~2.1 by 11.9 points and multi-round performance on EvoCode-Bench v2 MT@4 by 13.8 points.
\end{sloppypar}
\end{abstract}

\section{Introduction}
\label{sec:intro}

As terminal-based code agents become increasingly prevalent, the trajectories they produce have accumulated at scale, while realistic, executable environments remain scarce.
This gap matters because a trajectory and an environment are not equally useful. 
A trajectory is a single, fixed record: its quality is bounded by the responses of the policy model that produced it, and we cannot check whether the changes it made to the codebase are actually correct. 
An environment has none of these limits, because the same task can be solved again by a stronger model, the result can be verified by our own tests, and harder tasks can be posed on the same workspace. 
For post-training, the environment is therefore the resource worth scaling.
Expert-authored benchmarks show what a good environment looks like. Terminal-Bench~\citep{terminalbench2026}, for example, pairs every task with a custom container, an instruction, and an executable verifier. 
This reliability comes from manual effort, which also limits how many tasks can be built, while effective post-training needs far more environments than manual curation can provide. 
Supplying realistic, verifiable environments at training scale is therefore the challenge we address.

Existing work scales executable environments in three ways.
\textbf{Repository-based methods} build tasks from the git history of a real repository~\citep{pan2024swegym,jain2025r2egym}.
For a bug that was fixed in the past, they use git to roll the repository back to the state just before the fix to form the environment, take the original bug report as the task, and reuse the tests that came with the fix to check the solution.
\textbf{Perturbation methods} take a repository that passes its tests, inject a bug so that some tests now fail, and ask the agent to fix it~\citep{swesmith2025,cligym2026}. 
A few repositories can yield many tasks this way, but every task is a repair task, and the range of tasks is limited by the source repositories and the kinds of bugs that can be injected.
\textbf{Task-conditioned synthesis} generates the task and its environment together from scratch, guided by categories or compositional axes~\citep{endlessterminals2026,tmax2026}, skill taxonomies~\citep{hua2026cliuniverseverifiabletasksynthesis}, or skill graph~\citep{fan2026skillsynth}. 
These methods offer control over coverage and task-environment alignment, but the environment is not tied to any real project, so its realism depends entirely on the generator, which tends to produce small and tidy workspaces rather than real code.

\begin{figure}[t]
\centering
\includegraphics[width=\textwidth]{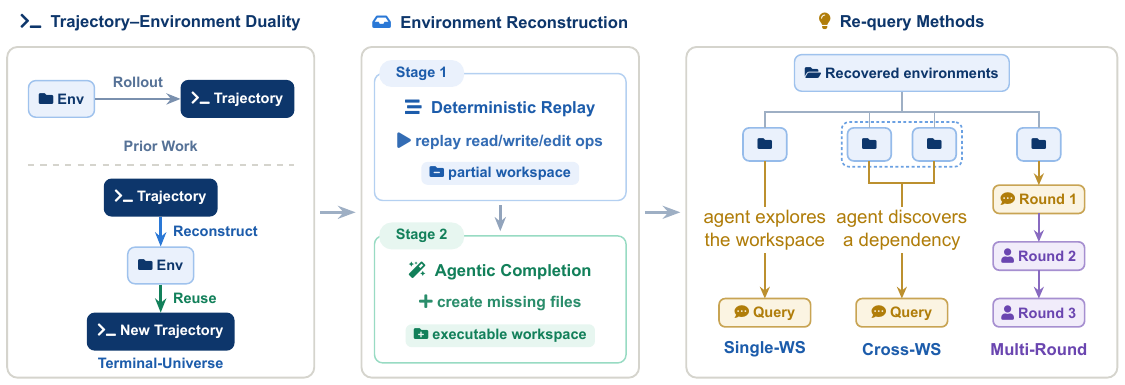}
\caption{\textbf{Overview of Terminal-Universe.} A trajectory and an environment are two views of the same episode, so prior work rolls out a trajectory from an environment while we invert the mapping and recover an environment from a trajectory (left). Reconstruction proceeds in two stages, deterministic replay followed by agentic completion (middle). Each recovered environment is then re-queried within one workspace, across multiple dependent workspaces, or over multiple user rounds (right). \Cref{fig:framework} details each mechanism.}
\label{fig:overview}
\end{figure}

Across these routes, environment construction either starts from an existing environment or is coupled to a newly generated task from scratch.
Trajectories, as observations of the environments they ran in, remain largely underexplored as a resource.
The tool calls recorded inside a trajectory already reveal the content and structure of the environment it ran in. 
For instance, the tool \texttt{Read} shows file contents, and \texttt{Write} and \texttt{Edit} show how the workspace changed. 
This is enough to rebuild an executable copy of the original workspace. 
Therefore, we propose \textbf{Terminal-Universe} (\Cref{fig:overview}), a framework which turns each trajectory into a reusable environment and explores it for synthesizing new tasks and continued interactions.

Specifically, to construct the environments based on the trajectories (\Cref{sec:reconstruction}), we first replay the recorded file operations and restore each file to its state before the agent changed it, holding back the agent's own edits so that the workspace starts unsolved. 
Since replay usually yields only a partial workspace, a completion agent then automatically supplies the missing files and dependencies the task needs, without leaking the solution.
For the corresponding tasks, we both reconstruct the original intent queries and synthesize entirely new ones. 
Besides, to better match real-world software engineering scenarios, we also scale the tasks on the environment along two complementary axes: breadth and depth (\Cref{sec:query}).
\textbf{Breadth expansion} goes beyond the single-repository setting that most task synthesis assumes. 
We identify dependency relations between related environments and build cross-workspace tasks that span multiple codebases which are much harder, e.g., reading a reference implementation, moving a feature from one project to another, or connecting two components. 
\textbf{Depth expansion} turns a single-round task into a multi-round one. 
After the agent completes an initial query, a user agent asks grounded follow-ups as the workspace changes—extending it with new requirements, or, when a round fails, asking for a fix based on what went wrong. We verify every round, and the outcome is fed back to the agent as natural, user-visible feedback.
Every new task comes with a verifier written by an agent inside the container, and we keep only the trajectories whose tests all pass (\Cref{sec:scaling}). 

On publicly available terminal agent trajectories, this pipeline produces $37.3$k task-sufficient environments. 
Over these environments, we generate new queries and use Qwen3.7-Max as the teacher to roll out a solution trajectory for each task. 
Training Qwen3.5-27B on the resulting data validates the effectiveness of our method: it improves the single-round benchmark Terminal-Bench~2.1 by $11.9$ points and the multi-round benchmark EvoCode-Bench v2 MT@4 by $13.8$ points (\Cref{sec:experiments}).
Extensive ablations confirm that each component contributes, most notably that re-solving tasks in reconstructed environments far outperforms imitating the raw trajectories.

\noindent To sum up, we make the following contributions:
\begin{itemize}[leftmargin=1.4em,itemsep=2pt,topsep=2pt]
  \item \textbf{Environments reconstruction.} We reframe recorded agent trajectories as a source of reusable executable environments, and reconstruct each one through deterministic replay followed by agentic completion---without needing the original repository or building from scratch.
  \item \textbf{Re-querying methods.} We scale the utility of each reconstructed environment through new task generation. Besides the vanilla single-workspace and single-turn tasks, we contribute breadth expansion via cross-workspace task synthesis and depth expansion via multi-round user queries. Each task is paired with an agent-authored verifier, and only trajectories that pass are kept.
  \item \textbf{Empirical validation.} Fine-tuning Qwen3.5-27B on the resulting corpus improves Terminal-Bench~2.1 by $11.9$ points and EvoCode-Bench v2 MT@4 by $13.8$ points over the same base model. Ablations confirm that each component contributes, and notably that re-solving in reconstructed environments far outperforms imitating the raw trajectories.
\end{itemize}

\begin{table}[t]
\centering
\caption{Comparison of representative environment and training-data construction methods. Strategy summarizes the primary mechanism used to construct environments or training tasks. Env. counts reusable source repositories/workspaces or independently constructed environments, and Tasks gives the reported task or training-instance count. Verif. indicates an executable task verifier, Multi-Round denotes extended multi-round queries, and Cross-WS cross-workspace synthesis.}
\label{tab:positioning}
{\small
\setlength{\tabcolsep}{2.8pt}
\renewcommand{\arraystretch}{1.18}
\begin{tabularx}{\textwidth}{@{}>{\raggedright\arraybackslash}X *{6}{c}@{}}
\toprule
\textbf{Method} & \textbf{Strategy} & \textbf{Env.} & \textbf{Tasks} & \textbf{Verif.} & \textbf{Multi-Round} & \textbf{Cross-WS} \\
\midrule
Endless Term.~\citep{endlessterminals2026} & task-conditioned          & 3{,}255 & 3{,}255 & \tyes & \tno & \tno \\
\rowcolor{tblzebra} TMax~\citep{tmax2026}  & combinatorial sampling    & 14.6k   & 14.6k   & \tyes & \tno & \tno \\
CLI-Gym~\citep{cligym2026}             & env. perturbation         & 29      & 1{,}655 & \tyes & \tno & \tno \\
\rowcolor{tblzebra} CLI-Universe~\citep{hua2026cliuniverseverifiabletasksynthesis} & taxonomy-guided & 6k & 6k & \tyes & \tno & \tno \\
SkillSynth~\citep{fan2026skillsynth}   & skill graph-guided        & 3{,}560 & 3{,}560 & \tyes & \tno & \tno \\
\rowcolor{tblzebra} OpenThinker-Agent~\citep{otagent2026} & multi-source curation & --      & 100k    & \tyes & \tno & \tno \\
RST~\citep{li2026recursivesynthesislonghorizonterminal} & recursive evolution & 37.5k & 37.5k & \tyes & \tno & \tno \\
\rowcolor{tblzebra} CalibForge~\citep{meng2026calibforge} & adversarial calibration & 5{,}431 & 5{,}431 & \tyes & \tno & \tno \\
\midrule
\rowcolor{promptcardgreenbg}
Terminal-Universe (ours)                & trajectory reconstruction & 37.3k & 32.0k & \tyes & \tyes & \tyes \\
\bottomrule
\end{tabularx}
}
\end{table}

\section{Related Work}
\label{sec:related}

\textbf{Environment scaling for terminal agents.} Existing methods scale executable workspaces through three main routes: recovering repository states from real development histories, modifying functioning workspaces, or constructing task-specific environments from generated specifications. SWE-Gym~\citep{pan2024swegym} and R2E-Gym~\citep{jain2025r2egym} construct environments from repository versions associated with historical issues or commits. SWE-smith~\citep{swesmith2025} and CLI-Gym~\citep{cligym2026} start from working repositories or CLI workspaces and introduce bugs or failures into them. SETA~\citep{shen2026seta} instead scales verifiable terminal environments for reinforcement learning through synthesis and adaptive environment evolution. A separate family jointly generates terminal tasks and their containerized execution environments; we discuss their task-construction strategies below. Terminal-Universe instead takes recorded agent trajectories as its starting point. It replays the file operations recorded in each trajectory and uses agentic completion to restore missing project context, producing executable workspaces that can be reused.

\textbf{Terminal task synthesis.} Existing methods derive tasks either from abstract priors or concrete executable environments. Top-down pipelines generate task-environment pairs from categories or seeds~\citep{endlessterminals2026,nemotronterminal2026,termigen2026}, while more structured variants organize generation around capability taxonomies, compositional axes, or skill graphs~\citep{hua2026cliuniverseverifiabletasksynthesis,tmax2026,fan2026skillsynth}. Other methods use agent skills, executable meta-tasks, or solver feedback to diversify and calibrate the generated tasks~\citep{cheng2026terminalworld,pan2026metatask,meng2026calibforge}.

Environment-grounded methods instead derive tasks from working code and project context. They perturb healthy environments, mine repository documentation, ground tasks in real issues, or jointly realize instructions, solutions, and verifiers in a shared environment~\citep{cligym2026,terminaltraj2026,yang2026interactiontrajectories,shi2026facet}. RST recursively expands verified seeds by lengthening solutions and realigning their tasks, verifiers, and environments~\citep{li2026recursivesynthesislonghorizonterminal}. Terminal-Universe takes past agent trajectories as its starting point: it reconstructs their workspaces before generating new tasks within or across them.

\textbf{Interactive and multi-round agents.} Moving beyond isolated, single-turn prompts, recent benchmarks model agent interaction as dynamic, multi-turn dialogues. InterCode formalizes interactive coding via execution feedback in containerized shells~\citep{yang2023intercode}. SWE-INTERACT~\citep{raghavendra2026sweinteractreimaginingswebenchmarks} uses a simulated user to progressively reveal requirements and provide targeted revisions, while SWE-Together~\citep{wu2026swetogetherevaluatingcodingagents} replays real user--agent sessions through a state-conditional user simulator that provides feedback as an evaluated agent progresses. ICAE-Bench~\citep{peng2026icaebenchevaluatingcodingagents} evaluates interactive project construction from incomplete product requirements using an automated user agent. In persistent workspaces, EvoCode-Bench v2~\citep{shen2026evocodebench} evaluates agents across sequential development requests. Terminal-Universe embraces this interactive setting through depth expansion, extending reconstructed workspaces into multi-round sessions that capture iterative user feedback and requirement refinement.

\Cref{tab:positioning} positions Terminal-Universe against representative methods. In contrast to prior work, it builds its environments from recorded trajectories and scales tasks along both breadth (cross-workspace synthesis) and depth (multi-round queries).

\section{Terminal-Universe}
\label{sec:method}

\begin{figure*}[t]
\centering
\includegraphics[width=\textwidth]{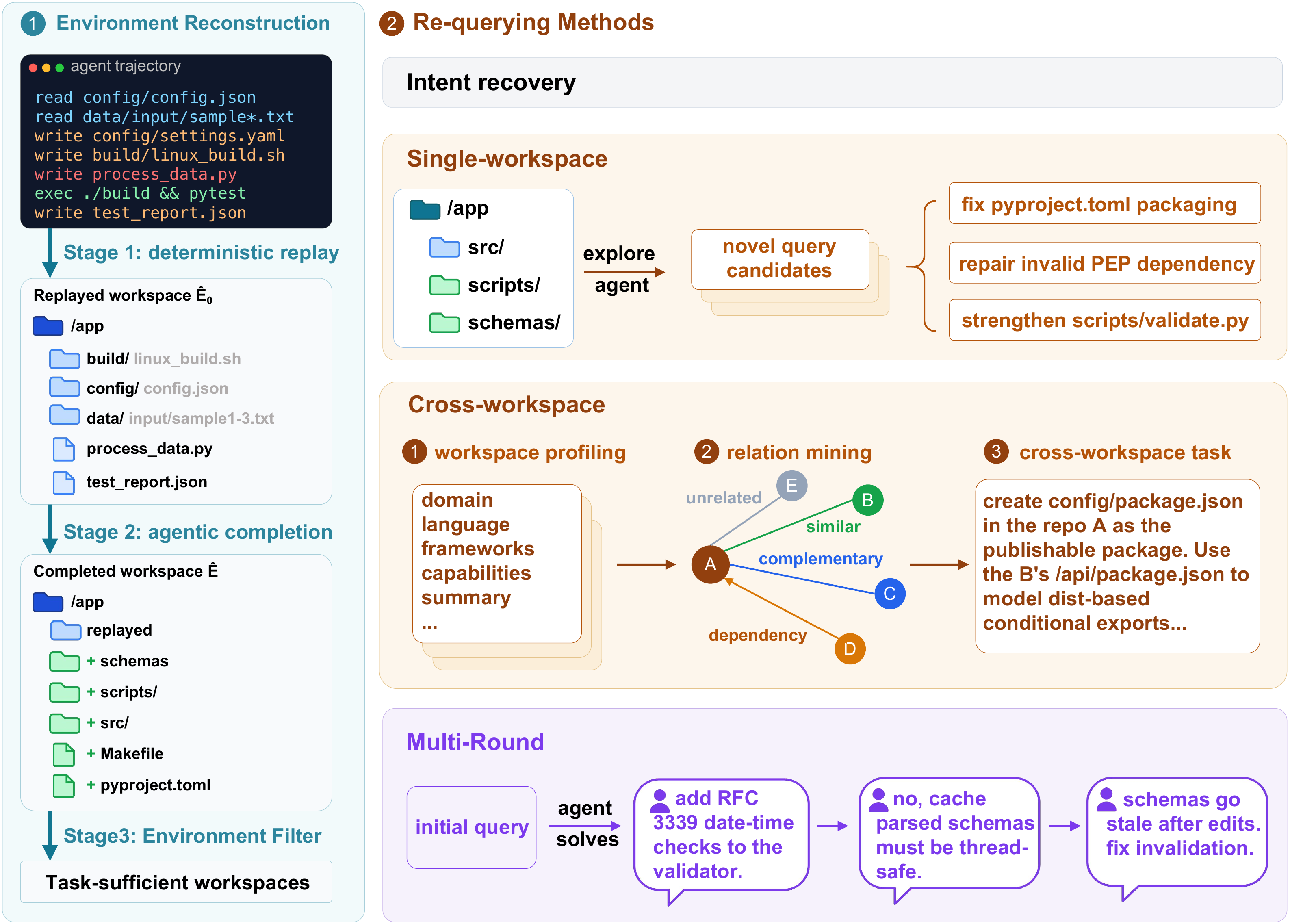}
\caption{\textbf{Framework of Terminal-Universe.} A recorded trajectory is replayed and completed into an executable workspace, which is retained only when it is sufficient for its task (left). The recovered workspace supports Intent Recovery, Single-WS synthesis that explores a workspace to propose novel task candidates, Cross-WS synthesis via workspace profiling and relation mining, and Multi-Round continuation that extends an initial query into an iterative session.}
\label{fig:framework}
\end{figure*}

\Cref{fig:framework} details the mechanisms across environment reconstruction, the four re-querying variants, and verification. The following subsections describe environment reconstruction (\Cref{sec:reconstruction}), re-querying (\Cref{sec:query}), and verification (\Cref{sec:verification}), respectively.

\subsection{Environment Reconstruction}
\label{sec:duality}
\label{sec:reconstruction}

From the file and command operations recorded in a trajectory $\tau$, we recover an executable workspace $\widehat E$ that approximates the environment $E$ in which $\tau$ was produced. This recovery is inherently lossy, since unaccessed files, implicit system dependencies, and external network resources leave no direct trace. We therefore reconstruct in three stages: deterministic replay recovers the file states $\tau$ exposes directly, agentic completion supplies the latent context it omits, and environment filtering keeps only workspaces sufficient for the recovered task $q$.

\textbf{Stage 1: deterministic replay.} Replay processes the read, write, and edit operations in $\tau$ in chronological order to recover, for each accessed path, the earliest and latest file contents visible in the trajectory. The reconstructed initial workspace $\widehat E_0$ collects each pre-existing file at its earliest observed version, before the agent's first change; files created by the agent are excluded, and the agent's file changes are recorded separately for later verification. Since the trajectory exposes only the paths the agent touched, and file contents may be incomplete when the trajectory shows only part of a file or terminal output is truncated, $\widehat E_0$ remains a partial workspace.

\textbf{Stage 2: agentic completion.} Given the partial workspace $\widehat E_0$ and the recovered task $q$, a completion agent creates missing files, completes partial files, and restores dependencies needed to make $q$ solvable without implementing it. We denote the resulting completed workspace by $\widehat E$. We apply this stage to all replayed workspaces, and its effect on workspace complexity is detailed in \Cref{sec:scaling}.

\textbf{Stage 3: environment filtering.} A completed workspace is useful only when it exposes enough project context to support its task. An agentic judge inspects each $\widehat E$ with read-only shell and file tools and, conditioned on the recovered task $q$, labels the workspace sufficient or insufficient according to whether its source, configuration, data, and structure give a capable agent enough context to work on the task (\Cref{app:completeness-prompt}). Only sufficient workspaces are retained for downstream task generation, and the per-stage sufficiency rates and final counts are reported in \Cref{sec:scaling}.

Each reconstructed workspace runs in a standardized \texttt{ubuntu:24.04} container with network access. Compared with repository-specific images, this design lowers cost and simplifies deployment, with a modest reduction in resolve rate reported by prior work~\citep{zeng2026dockerlessenvironmentfreeprogramverifier}.

\subsection{Re-querying}
\label{sec:query}

While reconstruction yields executable environments, re-querying dictates how effectively their latent capability space is exploited. We introduce four complementary re-querying mechanisms, denoted throughout the paper as Intent Recovery, Single-WS, Cross-WS, and Multi-Round. Intent Recovery reconstructs source tasks, while Single-WS synthesizes new tasks within individual workspaces. Cross-WS provides breadth by connecting related workspaces, and Multi-Round provides depth by extending an initial query into an interactive session.

\begin{mechheader}
\textbf{Intent recovery.} Consolidate one or more source user requests into a self-contained task.
\end{mechheader}
\noindent We normalize each source trajectory into a chronological stream of user requests, agent actions, and file changes. For a single-round trajectory, the sole substantive user request directly defines the task. For a multi-round trajectory, the first substantive user request establishes the task topic, while later requests are incorporated when they clarify, constrain, or extend the same task; unrelated task shifts are excluded. We use agent actions and file evidence to interpret the requests, but retain only requirements stated by the user (\Cref{app:q1-prompt}).

\begin{mechheader}
\textbf{Single-workspace synthesis.} Derive new queries from individual reconstructed workspaces.
\end{mechheader}
\noindent An offline generator inspects each workspace and synthesizes five self-contained candidate tasks under groundedness, structural diversity, and verifiability constraints. We randomly select one valid candidate per environment for rollout and verification.

\smallskip
\begin{mechheader}
\textbf{Breadth expansion.} Connect related workspaces in a single query.
\end{mechheader}
\noindent\textbf{Cross-workspace synthesis.} To synthesize tasks spanning multiple codebases, we discover directional dependency relationships across recovered environments. An agent first profiles each workspace's technical domain and implemented capabilities. We then retrieve candidate pairs via TF-IDF nearest-neighbor search, and employ an LLM judge to identify directional dependency edges where a target workspace lacks a capability already implemented in a reference workspace.

We allocate exactly one task per pair. Each cross-workspace task pairs a writable target workspace with a read-only reference workspace mounted at a separate path. The task generator confirms that the functional gap is genuine and specifies observable behaviors in the target that bridge this gap, verifiable via deterministic local commands. The query provides only these target behaviors and the reference's mount path, not its internal details, keeping the reference a genuine dependency. The solver must therefore navigate, internalize, and adapt the reference implementation on its own.

\begin{mechheader}
\textbf{Depth expansion.} Extend solved terminal tasks through user-agent follow-ups.
\end{mechheader}
\noindent\textbf{Multi-round user queries.} Beginning with an initial terminal query, we preserve the workspace across coding rounds and introduce a user agent after the initial response. The continuation process employs two coordinated mechanisms to ensure coherent and realistic multi-turn trajectories:

\begin{enumerate}[leftmargin=1.35em,itemsep=2pt,topsep=2pt]
\item \textbf{Evolving task specification.} The user agent maintains an explicit requirement tracker logging active, satisfied, and updated requirements. Prior to each follow-up, it updates this tracker by adding, modifying, or replacing constraints, maintaining contextual coherence as the workspace evolves.

\item \textbf{Round-level verification and feedback.} At each extension round, the user agent commits the updated specification, prompting an automated verifier to author round-level acceptance tests before the coding agent acts. Upon receiving the agent's response, the test runner evaluates both new criteria and active regression checks. The solver agent is strictly isolated from test scripts and tracebacks; the user agent interprets the structured test results and translates any failure into natural, user-observable complaints. Intermediate failures are retained within the conversation history, providing realistic supervision for error diagnosis and recovery.
\end{enumerate}

\begin{wrapfigure}[20]{r}{0.53\textwidth}
\centering
\vspace{-2.8\baselineskip}
\includegraphics[width=0.385\textwidth]{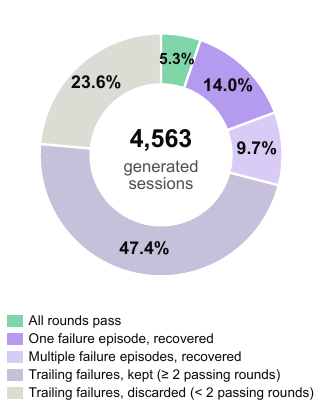}
\caption{Pass/fail patterns in Multi-Round sessions.}
\label{fig:q4-rounds-shapes}
\vspace{-0.8\baselineskip}
\end{wrapfigure}
Across rounds, the user agent's requests fall into three interaction styles: (i) feature extension, which introduces new grounded requirements following successful verification; (ii) feature revision, triggered by test failures to demand bug fixes based on observed behavior; and (iii) feature conflict, which modifies or overrides prior specifications to reflect changing user intent. The style of each request follows naturally from the round outcome and the session context, and the resulting distribution is reported as observed (\Cref{app:user-continuation-case-study}, \Cref{tab:q4-style-distribution}). Sessions continue for up to six follow-up rounds or until a termination sentinel is emitted.

\Cref{fig:q4-rounds-shapes} reports pass/fail patterns for the continuation data after the round-level selection of \Cref{sec:verification}: the $3{,}079$ retained records average $4.51$ rounds, and $69.6\%$ contain a failure that the session subsequently repairs, preserving recovery supervision.

\subsection{Verification and Filtering}
\label{sec:rollout}
\label{sec:verification}

\noindent\textbf{Agentic verifier construction.}
Each Single-WS or Cross-WS task is paired with an executable verifier authored by a dedicated agent within the target container. Conditioning on the task specification and workspace files, the verifier agent crafts a self-contained pytest suite through iterative local execution, strictly evaluating the public interfaces and expected behaviors specified in the prompt (\Cref{app:verification}).

\noindent\textbf{Solution rollout.} Candidate solutions are rolled out using the teacher model\footnote{All model-driven components in this work use Qwen3.7-Max (xhigh effort) as the teacher model.} within the Claude Code scaffold~\citep{anthropic2026claudecode} inside the task container. Rollouts decode with temperature $1.0$, top-$p=0.95$, and interleaved thinking over a $256$k-token context window, with turn responses capped at $65{,}536$ tokens, proactive summarization triggered at $176$k tokens, and a maximum of $500$ agent turns under a four-hour wall-clock timeout. The task environment provides containerized network access to fetch absent dependencies (\Cref{sec:reconstruction}).

\noindent\textbf{Verification and data selection.} For Single-WS and Cross-WS, the authored test suite runs against the final workspace state after rollout, and a trajectory is accepted only if all tests pass. Multi-Round is selected at the round level instead: we trim trailing suffixes of consecutive failed rounds from terminated sessions and retain a session only if it contains at least two verified passing rounds. Intermediate failures that precede a successful recovery are kept, supplying supervision for error diagnosis and recovery. Accepted trajectories are formatted into multi-turn SFT demonstrations and undergo strict decontamination against evaluation benchmarks (\Cref{sec:experiments}).

To provide concrete intuitions for the mechanisms above, we detail several comprehensive case studies in the Appendix. \Cref{app:case-study} traces a trajectory-to-query example, \Cref{app:cross-workspace-case-study} follows a cross-workspace repository set from candidate sourcing to a selected task, and \Cref{app:user-continuation-case-study} follows a multi-round continuation from verifier-driven revision to controlled requirement change (\Cref{fig:q4-trajectory}). \Cref{sec:scaling} reports dataset sizes, and \Cref{app:sftstats} reports interaction statistics.

\section{Data Construction at Scale}
\label{sec:scaling}

\subsection{Seed Selection}

We source raw agent trajectories from diverse terminal-style CLI and software-engineering corpora. An execution trace is retained as a viable seed if the workspace state observed at its end contains at least $5$ files and $100$ lines. To prevent data leakage, we strictly filter out all Terminal-Bench-derived sources. Complete source breakdowns and selection statistics are detailed in \Cref{app:sources}.

\subsection{Reconstruction Statistics}

\begin{figure}[H]
\centering
\includegraphics[width=\textwidth]{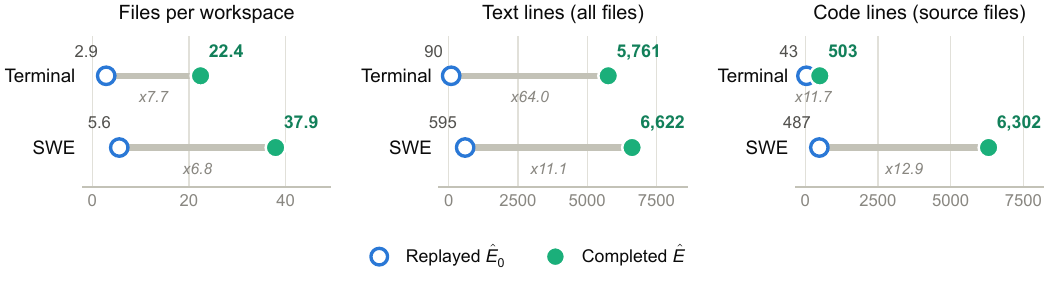}
\caption{Mean workspace size before and after agentic completion.}
\label{fig:complexity}
\end{figure}

From the source trajectory pool, the reconstruction pipeline yields $68{,}263$ reconstructed environments. As illustrated in \Cref{fig:complexity}, replayed terminal workspaces initially contain few files because solution files generated during the original rollout are withheld. Agentic completion substantially enriches these environments, increasing the mean file count from $2.9$ to $22.4$. While SWE seeds originate from richer repository states, completion similarly expands their contextual breadth (full complexity distributions appear in \Cref{tab:complexity}).

After contamination filtering and repository-level deduplication, we apply the Stage 3 sufficiency judge (\Cref{sec:reconstruction}) to the surviving completed environments.

\begin{table}[H]
\centering
\captionsetup{font=small}
\small
\caption{Workspace sufficiency under recovered intents.}
\label{tab:workspace-completeness-census}
\setlength{\tabcolsep}{3pt}
\renewcommand{\arraystretch}{1.18}
\begin{tabularx}{\textwidth}{@{}l c|*{2}{>{\centering\arraybackslash}X}@{}}
\toprule
\multirow{2}{*}{\textbf{Pool}} & \multirow{2}{*}{\textbf{\# Environments}} & \multicolumn{2}{c}{\textbf{Sufficient (\%)}} \\
\cmidrule(l){3-4}
 & & \textbf{Post-replay} & \textbf{Post-completion} \\
\midrule
Terminal & 38{,}294 & 40.2 & \tbest{93.5} \\
SWE & 1{,}900 & 20.1 & \tbest{77.1} \\
\bottomrule
\end{tabularx}
\end{table}

As shown in \Cref{tab:workspace-completeness-census}, deterministic replay alone yields a sufficiency rate of $40.2\%$ for Terminal and $20.1\%$ for SWE workspaces. Agentic completion significantly improves this coverage to $93.5\%$ and $77.1\%$, respectively, across $38{,}294$ evaluated terminal environments and $1{,}900$ SWE repositories (one representative reconstruction each) after repository-level deduplication. In total, task-grounded assessment identifies $37{,}273$ fully sufficient environments suitable for downstream task generation (\Cref{app:sources}).

\begin{figure}[H]
\centering
\includegraphics[width=0.98\linewidth]{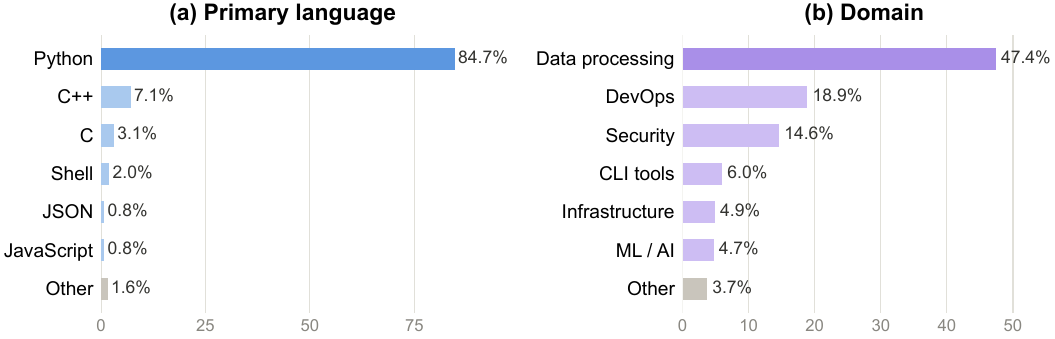}
\caption{Language and domain composition of reconstructed terminal workspaces.}
\label{fig:env-composition}
\end{figure}

\Cref{fig:env-composition} illustrates the diversity of the reconstructed terminal pool: Python represents the dominant primary language ($84.7\%$), alongside C++ and C, with data processing, DevOps, and security workloads collectively accounting for over $80\%$ of technical domains.

\subsection{SFT Corpus Composition}

These task-sufficient environments serve as the foundation for our four re-querying variants: Intent Recovery reconstructs the original tasks, Single-WS yields novel within-repository tasks, Cross-WS discovers inter-repository dependency challenges, and Multi-Round extends initial queries into iterative sessions. Across the terminal pool, verifier-filtered Single-WS, Cross-WS, and Multi-Round synthesis yields $31{,}977$ SFT demonstrations ($25{,}386$ Single-WS, $3{,}512$ Cross-WS, and $3{,}079$ Multi-Round trajectories), totaling approximately $1.42$B training tokens. Median interaction statistics are provided in \Cref{app:sftstats}.

\section{Experiments}
\label{sec:experiments}

\subsection{Setup}

\textbf{Training.} We fine-tune Qwen3.5-27B by SFT on Terminal-Universe data for two epochs. We use a constant learning rate of $7\times10^{-6}$, a global batch size of $256$, and a $256$k-token sequence length. Before training, we run a 13-gram contamination check against Terminal-Bench tasks and exclude Terminal-Bench-derived source datasets from all four re-querying variants.

{\setlength{\emergencystretch}{2em}
\textbf{Single-round evaluation (Terminal-Bench 2.0 and 2.1).} We evaluate Terminal-Bench with Claude Code (version 2.1.126) and Terminus2~\citep{terminalbench2026} using the XML parser. Evaluation mirrors the decoding and execution configurations of the solution rollout (\Cref{sec:rollout}), employing temperature $1.0$, top-$p=0.95$, interleaved thinking, a $256$k context window, a $65{,}536$-token turn limit, proactive summarization at $176$k tokens, up to $500$ agent turns, and a four-hour wall-clock timeout. For Claude Code, interactive and web-retrieval tools (\texttt{WebFetch}, \texttt{WebSearch}, \texttt{AskUserQuestion}, \texttt{EnterPlanMode}, \texttt{ExitPlanMode}) are disabled. Each container runs with 12 CPU cores and 32 GiB of memory. Reported scores represent mean pass rates over six independent runs. Differences between Terminal-Bench 2.0 and 2.1 are documented in the benchmark update.\footnote{\url{https://github.com/harbor-framework/terminal-bench-2/pull/53}}\par}

\textbf{Multi-round evaluation (EvoCode-Bench v2).} We evaluate depth expansion on EvoCode-Bench v2\footnote{\url{https://unipat.ai/benchmarks/EvoCode-Bench}}~\citep{shen2026evocodebench}, which contains 26 coding tasks and 227 rounds (5--15 requests per task). Workspaces and sessions persist, and cumulative verifiers check both current and active prior requirements. Our Qwen3.5-27B checkpoints use Terminus2-XML at temperature $1.0$ with interleaved thinking, a $256$k-token context, a per-turn limit of $65{,}536$ tokens, proactive summarization at $176$k tokens, up to $500$ agent turns per request round, and a total wall-clock limit of $10$ hours per stateful task. Each task is evaluated in four independent runs. MT@4 uses fail-stop scoring: a run receives credit only for the consecutive rounds it passes before its first failure, and each task-round is credited if any run reaches it successfully. The score averages credited rounds within each task and then across tasks. Case score averages verifier-case pass rates within tasks and then across tasks and runs to measure partial progress.

\subsection{Main Results}
\label{sec:main-results}

{\setlength{\emergencystretch}{1em}
\Cref{tab:main} reports single- and multi-round performance. Fine-tuning Qwen3.5-27B on the Full Mixture (the verifier-filtered Single-WS, Cross-WS, and Multi-Round trajectories, $32.0\text{k}$ records) reaches $52.8\%$ on Terminal-Bench 2.0 and $58.1\%$ on Terminal-Bench 2.1 ($+11.2$ and $+11.9$ over base, respectively) under Terminus2-XML. Under Claude Code, it reaches $58.2\%$ on Terminal-Bench 2.1 ($+10.4$ over base).\par}

\begin{table}[H]
\centering
\caption{Terminal-Bench and EvoCode-Bench v2 results (\%). Terminal-Bench columns report Avg.\ Pass@1. \tbest{Green bold} marks the best score among task synthesis methods, \tsecond{blue underline} the second best. $^{*}$ marks scores we measure on the released models under the configurations in \Cref{app:openweights-eval-config}; other scores are taken from the corresponding reports.}
\label{tab:main}
{\scriptsize
\setlength{\tabcolsep}{2pt}
\renewcommand{\arraystretch}{1.18}
\begin{tabularx}{\textwidth}{@{}>{\raggedright\arraybackslash}X ccc|cc|cc@{}}
\toprule
\multirow{2}{*}{\textbf{Model}} & \multirow{2}{*}{\textbf{Base Model}} & \multirow{2}{*}{\textbf{Data size}} & \multirow{2}{*}{\makecell{\textbf{TB}\\\textbf{Scaffold}}} & \multicolumn{2}{c|}{\textbf{Terminal-Bench}} & \multicolumn{2}{c}{\textbf{EvoCode-Bench v2}} \\
\cmidrule(lr){5-6}\cmidrule(l){7-8}
 & & & & \textbf{TB2.0} & \textbf{TB2.1} & \textbf{MT@4} & \textbf{Case Score} \\
\midrule
\rowcolor{tblbandgray}
\tblock{8}{promptcardgray}{Base and teacher models} \\
Qwen3.5-27B & -- & -- & Terminus2 & 41.6 & 46.2 & $6.3^{*}$ & $67.8^{*}$ \\
\rowcolor{tblzebra} Qwen3.7-Max & -- & -- & Terminus2 & 69.7 & 74.5 & $39.8^{*}$ & $83.4^{*}$ \\
\midrule
\rowcolor{promptcardbluebg}
\tblock{8}{promptcardbluedark}{Terminal task synthesis methods} \\
TerminalTraj-32B~\citep{terminaltraj2026} & Qwen2.5-Coder-32B & 50.7k & Terminus2 & 22.0 & $28.5^{*}$ & $0.0^{*}$ & $15.9^{*}$ \\
\rowcolor{tblzebra} TermiGen-32B~\citep{termigen2026} & Qwen2.5-Coder-32B & 3.3k & BashAgent & 19.3 & $21.3^{*}$ & $0.0^{*}$ & $12.2^{*}$ \\
LiberCoder-235B~\citep{cligym2026} & Qwen3-235B-A22B & 48.3k & OpenHands & 31.0 & -- & -- & -- \\
\rowcolor{tblzebra} Nemotron-Terminal-32B~\citep{nemotronterminal2026} & Qwen3-32B & 490k & Terminus2 & 27.4 & $27.9^{*}$ & $0.0^{*}$ & $13.7^{*}$ \\
SkillSynth-32B~\citep{fan2026skillsynth} & Qwen3-32B & 10.7k & Terminus2 & 29.6 & -- & -- & -- \\
\rowcolor{tblzebra} Terminal-World-32B~\citep{cheng2026terminalworld} & Qwen3-32B & 5.7k & Terminus2 & 31.5 & -- & -- & -- \\
Terminal-Lego-32B~\citep{yang2026interactiontrajectories} & Qwen3-32B & 15.3k & Terminus2 & 24.3 & -- & -- & -- \\
\rowcolor{tblzebra} CLI-Universe-32B~\citep{hua2026cliuniverseverifiabletasksynthesis} & Qwen3-32B & 6k & Terminus2 & 33.4 & -- & -- & -- \\
TMax-27B~\citep{tmax2026} & Qwen3.6-27B & 14.6k & Vanillux2Agent & 42.7 & 44.9 & \tsecond{$17.1^{*}$} & \tsecond{$72.5^{*}$} \\
\rowcolor{tblzebra} OpenThinker-Agent-32B~\citep{otagent2026} & Qwen3-32B & 100k & Terminus2 & 26.2 & $30.7^{*}$ & $0.0^{*}$ & $19.7^{*}$ \\
Meta-Task-32B~\citep{pan2026metatask} & Qwen3-32B & 3.2k & Terminus2 & 31.8 & -- & -- & -- \\
\rowcolor{tblzebra} RST-27B~\citep{li2026recursivesynthesislonghorizonterminal} & Qwen3.5-27B & 37.5k & Terminus2 & \tsecond{49.4} & -- & -- & -- \\
CalibForge-35B-A3B~\citep{meng2026calibforge} & Qwen3.5-35B-A3B & 5.4k & CalibForge-Eval & 47.6 & -- & -- & -- \\
\rowcolor{tblzebra} FACET-Terminal-27B~\citep{shi2026facet} & Qwen3.5-27B & 1.2k & Terminus2 & -- & \tsecond{47.6} & -- & -- \\
\midrule
\rowcolor{promptcardgreenbg}
Terminal-Universe-27B & Qwen3.5-27B & 32.0k & Terminus2 & \tbest{52.8} & \tbest{58.1} & \tbest{20.1} & \tbest{76.1} \\
\bottomrule
\end{tabularx}
}
\end{table}

On EvoCode-Bench v2, the Full Mixture raises MT@4 from $6.3$ to $20.1$ and Case score from $67.8$ to $76.1$, showing that Full Mixture training also improves performance on persistent tasks with cumulative requirements.

\section{Analysis and Ablation Studies}
\label{sec:ablations}
Our method makes multiple design choices, and this section isolates the contribution of each. We ask: (1) is reconstructing an environment and re-solving it actually better than training on the original trajectory (\Cref{sec:resolving-analysis})? (2) does agentic completion matter, or is deterministic replay enough (\Cref{sec:completion-ablation})? (3) is verifier filtering worth the data it discards (\Cref{sec:verifier-ablation})? and (4) how should a fixed budget be spent across the re-querying axes—breadth (\Cref{sec:q2q3-profile}), depth (\Cref{sec:exp-depth}), and more environments versus more queries (\Cref{sec:ablation-expansion})? We finally test whether the pipeline transfers beyond terminal workspaces (\Cref{sec:cross-domain}).

\subsection{Task Re-solving}
\label{sec:resolving-analysis}

This ablation asks whether reconstruction is needed at all: is re-solving a recovered task better than simply imitating the original trajectory? \Cref{tab:task-resolving} compares the two on Terminal-Bench 2.1. To keep the comparison clean, Intent Recovery is not verifier-filtered, so verifier selection plays no role here.

\begin{table}[H]
\centering
\caption{Terminal-Bench 2.1 comparison of source-trajectory SFT and Intent Recovery (\%).}
\label{tab:task-resolving}
{\small
\setlength{\tabcolsep}{3pt}
\renewcommand{\arraystretch}{1.18}
\begin{tabularx}{0.72\textwidth}{@{}>{\raggedright\arraybackslash}X c c c c@{}}
\toprule
\textbf{Variant} & \textbf{Data size} & \textbf{Claude Code} & \textbf{Terminus2-XML} & \textbf{Avg.} \\
\midrule
Qwen3.5-27B & - & \score{47.8}{2.0} & \score{46.2}{3.9} & 47.0 \\
Source trajectories & 35.8k & \score{33.0}{3.9} & \score{40.3}{2.1} & 36.7 \\
\cmidrule(lr){1-5}
\rowcolor{promptcardgreenbg}
Intent Recovery & 35.8k & \bestscore{51.3}{3.1} & \bestscore{52.9}{1.4} & \tbest{52.1} \\
\bottomrule
\end{tabularx}
}
\end{table}

\textbf{Re-solving improves over source-trajectory SFT.} Source trajectories and Intent Recovery use the same Qwen3.5 chat template and tool-call schema for SFT. Source SFT preserves the original agents' behavior, whereas Intent Recovery uses a stronger teacher to re-solve recovered tasks in reconstructed workspaces. Source SFT averages $36.7$, compared with $52.1$ for Intent Recovery, suggesting that regenerating demonstrations under a consistent teacher policy provides more effective supervision.

\subsection{Impact of Agentic Completion}
\label{sec:completion-ablation}

This ablation asks whether agentic completion earns its cost, or whether deterministic replay alone is enough. We isolate the second reconstruction stage by training on Intent Recovery corpora whose environments are reconstructed with replay only, without completion. The replay-only variant trains on the $35{,}809$ source instances that deterministic replay reconstructs, and the completed variant is the production Intent Recovery corpus at the same volume. Both share the recovered queries and teacher configuration and differ only in the environment reconstruction mode.

\begin{table}[H]
\centering
\caption{Effect of agentic completion on Intent Recovery.}
\label{tab:completion-ablation}
\setlength{\tabcolsep}{9pt}
\renewcommand{\arraystretch}{1.18}
\begin{tabularx}{0.72\textwidth}{@{}>{\raggedright\arraybackslash}X c c@{}}
\toprule
\textbf{Variant} & \textbf{Records} & \textbf{Terminus2-XML} \\
\midrule
Qwen3.5-27B & -- & \score{46.2}{3.9} \\
\cmidrule(lr){1-3}
Replay only & 35.8k & \score{48.7}{3.5} \\
\rowcolor{promptcardgreenbg}
Replay + agentic completion & 35.8k & \bestscore{52.9}{1.4} \\
\bottomrule
\end{tabularx}
\end{table}

\textbf{Agentic completion helps.} At closely matched training volume, training on completed environments outperforms training on replay-only environments by $4.2$ points ($52.9$ versus $48.7$) and is more stable across runs ($\pm 1.4$ versus $\pm 3.5$). This is consistent with the sufficiency results in \Cref{tab:workspace-completeness-census}: replay alone leaves most terminal workspaces task-insufficient, and completion restores the execution context that re-solving depends on.

\textbf{Replay-only supervision still improves over the base model.} The replay-only variant still gains $2.5$ points over the base model ($46.2$ to $48.7$). Inspecting its rollouts reveals that, when presented with an incomplete workspace, the teacher often repairs missing files or setup before addressing the task. These trajectories remain useful, but some of the supervision focuses on repairing the workspace rather than solving the recovered task.

\subsection{Role of Verifier Filtering}
\label{sec:verifier-ablation}

This ablation asks whether discarding verifier-failed trajectories is worth the data it removes. For each synthesis method, we compare training on all teacher trajectories against training only on the verifier-passed subset from the same launched-task pool.

\begin{table}[H]
\centering
\caption{Effect of verifier filtering on Single-WS and Cross-WS.}
\label{tab:verifier-selection}
\setlength{\tabcolsep}{9pt}
\renewcommand{\arraystretch}{1.18}
\begin{tabularx}{0.72\textwidth}{@{}>{\raggedright\arraybackslash}X c c@{}}
\toprule
\textbf{Variant} & \textbf{Records} & \textbf{Terminus2-XML} \\
\midrule
Single-WS w/o verifier & 35.1k & \score{56.0}{3.3} \\
\rowcolor{promptcardgreenbg}
Single-WS w/ verifier & 25.4k & \bestscore{56.4}{2.6} \\
\cmidrule(lr){1-3}
Cross-WS w/o verifier & 7.1k & \score{53.2}{2.1} \\
\rowcolor{promptcardgreenbg}
Cross-WS w/ verifier & 3.5k & \bestscore{55.4}{2.7} \\
\bottomrule
\end{tabularx}
\end{table}

\textbf{Verifier filtering matters more for harder tasks.} For each synthesis method, we compare all teacher trajectories with the verifier-passed subset from the same launched-task pool. On Single-WS data, the verifier-passed subset performs similarly to the full corpus ($56.4$ versus $56.0$) with fewer records. On Cross-WS data, retaining verifier-failed trajectories lowers performance to $53.2$, whereas the verifier-passed subset reaches $55.4$ with less than half the data. The effect is also visible in Multi-Round synthesis: removing round-level verifier supervision lowers both EvoCode-Bench metrics (\Cref{tab:evocode-depth}).

\subsection{Breadth Expansion}
\label{sec:q2q3-profile}

This ablation asks whether cross-workspace tasks add supervision beyond single-workspace synthesis. \Cref{tab:breadth-results} evaluates Cross-WS against the Single-WS baseline. Cross-WS alone reaches $55.4$ under Terminus2-XML, while adding it to Single-WS raises performance from $56.4$ to $58.4$.

\begin{table}[H]
\centering
\caption{Terminal-Bench 2.1 results for Single-WS, Cross-WS, and their mixture (\%).}
\label{tab:breadth-results}
{\small
\setlength{\tabcolsep}{9pt}
\renewcommand{\arraystretch}{1.18}
\begin{tabularx}{0.72\textwidth}{@{}>{\raggedright\arraybackslash}X c c@{}}
\toprule
\textbf{Variant} & \textbf{Data size} & \textbf{Terminus2-XML} \\
\midrule
Qwen3.5-27B & -- & \score{46.2}{3.9} \\
\cmidrule(lr){1-3}
Single-WS & 25.4k & \score{56.4}{2.6} \\
Cross-WS & 3.5k & \score{55.4}{2.7} \\
\rowcolor{promptcardgreenbg}
Single-WS + Cross-WS & 28.9k & \bestscore{58.4}{2.1} \\
\bottomrule
\end{tabularx}
}
\end{table}

\begin{table}[H]
\centering
\caption{Trajectory and difficulty profiles of Single-WS and Cross-WS synthesis.}
\label{tab:q2q3-profile}
{\small
\setlength{\tabcolsep}{3pt}
\renewcommand{\arraystretch}{1.18}
\begin{tabularx}{\textwidth}{@{}l *{2}{>{\centering\arraybackslash}X}|*{2}{>{\centering\arraybackslash}X}|*{2}{>{\centering\arraybackslash}X}|c@{}}
\toprule
\multirow{2}{*}{\textbf{Variant}} & \multicolumn{2}{c|}{\textbf{Turns}} & \multicolumn{2}{c|}{\textbf{Tool calls}} & \multicolumn{2}{c|}{\textbf{Tokens / record}} & \multirow{2}{*}{\textbf{Teacher pass@1}} \\
\cmidrule(lr){2-3}\cmidrule(lr){4-5}\cmidrule(lr){6-7}
 & \textbf{Median} & \textbf{Mean} & \textbf{Median} & \textbf{Mean} & \textbf{Median} & \textbf{Mean} & \\
\midrule
Single-WS & 14 & 15.1 & 20 & 21.2 & 30.4k & 32.9k & $72.3\%$ \\
Cross-WS  & 23 & 25.3 & 38 & 39.7 & 46.5k & 51.9k & $49.2\%$ \\
\bottomrule
\end{tabularx}
}
\end{table}

\Cref{tab:q2q3-profile} contrasts the two synthesis variants on the same teacher. Cross-workspace tasks yield substantially longer and more complex trajectories: $1.6\times$ the assistant turns, $1.9\times$ the tool calls, and $1.5\times$ the tokens per record at the median. They are also harder for the teacher: pass@1 drops from $72.3\%$ on single-workspace tasks to $49.2\%$. Grounding the requirement in a second codebase forces the agent to read and reconcile two repositories before editing, which lengthens trajectories and raises task difficulty rather than merely adding data volume.

\begin{wrapfigure}[24]{r}{0.48\textwidth}
\centering
\vspace{-0.8\baselineskip}
\includegraphics[width=0.46\textwidth]{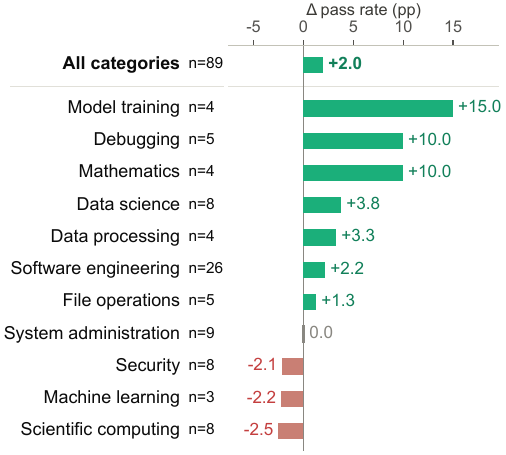}
\caption{Per-category Terminal-Bench 2.1 pass-rate changes when Cross-WS data is added to Single-WS training under Terminus2-XML. Single-task categories are omitted.}
\label{fig:q2q3-mix-categories}
\vspace{-0.8\baselineskip}
\end{wrapfigure}

\textbf{How Cross-WS expands breadth.} Single-WS provides the within-workspace baseline by synthesizing one novel task from each recovered environment. Cross-WS extends beyond this setting by requiring changes in a writable workspace grounded in evidence from a related read-only workspace. This exposes domain--operation combinations absent from single-workspace tasks and requires the agent to reconcile information across codebases.

\textbf{Where cross-workspace data helps.} Mixing single-workspace and cross-workspace records reaches $\score{58.4}{2.1}$ under Terminus2-XML, improving over the single-workspace baseline ($\score{56.4}{2.6}$). \Cref{fig:q2q3-mix-categories} breaks this comparison down by task category. Software engineering, the largest category with $26$ tasks, improves from $46.9$ to $49.1$, and larger gains appear in model training ($+15.0$) and debugging ($+10.0$). Three categories decline by $2.1$ to $2.5$ points, a range that falls within run-to-run variation for categories of at most eight tasks. The gain in this largest category, together with the aggregate improvement, suggests that cross-workspace data adds useful supervision beyond single-workspace synthesis.

\subsection{Multi-Round Depth Expansion}
\label{sec:exp-depth}

This ablation asks whether multi-round continuation data improves performance on persistent, cumulative tasks, and whether round-level verification during generation matters. We evaluate on EvoCode-Bench v2.

\textbf{Multi-round data improves depth.} Adding Multi-Round data to Single-WS raises MT@4 from $18.4$ to $21.0$ and Case score from $71.9$ to $76.9$. The higher MT@4 indicates that the model completes longer sequences of requests before its first failure, while the Case score gain shows that it passes a larger share of verifier cases across the session.

\textbf{Round-level verification directs effective continuation.} Within the matched Single-WS + Multi-Round comparison, removing round-level verifier feedback reduces MT@4 by $2.2$ points and Case score by $3.7$ points. Inspecting the resulting trajectories reveals that, without a grounded signal of what remains incorrect, continuation becomes less directed, yielding longer yet lower-quality multi-round trajectories.

\begin{table}[H]
\centering
\small
\caption{Effect of Multi-Round synthesis and round-level verification on EvoCode-Bench v2 (\%).}
\label{tab:evocode-depth}
\setlength{\tabcolsep}{7pt}
\renewcommand{\arraystretch}{1.18}
\begin{tabularx}{\textwidth}{@{}>{\raggedright\arraybackslash}X c|c c@{}}
\toprule
\textbf{Variant} & \textbf{Data size} & \textbf{MT@4 score} & \textbf{Case score} \\
\midrule
Single-WS & 25.4k & 18.4 & \score{71.9}{3.5} \\
\rowcolor{promptcardgreenbg}
Single-WS + Multi-Round & 28.5k & \tbest{21.0} & \bestscore{76.9}{1.4} \\
Single-WS + Multi-Round (w/o round verifier) & 28.5k & 18.8 & \score{73.2}{2.6} \\
\bottomrule
\end{tabularx}
\end{table}

\subsection{Expansion-Axis Ablation}
\label{sec:ablation-expansion}

This ablation asks where a fixed data budget is best spent. Starting from a common Single-WS base pool of $17{,}558$ environments with one query and one teacher solution each ($17.6$k records), we double the corpus along one axis at a time: adding a second set of environments, a second query per environment, or a second teacher solution per query. All three variants reach a matched budget of roughly $35$k records and are trained and evaluated under the setup of \Cref{sec:experiments} on Terminal-Bench 2.1, so any difference reflects how the budget is allocated rather than its size.

\begin{table}[H]
\centering
\caption{Data-scaling ablation for Single-WS synthesis. Starting from a common base pool, the corpus is doubled through additional environments, queries per environment, or solutions per query.}
\label{tab:expansion-axes}
{\small
\setlength{\tabcolsep}{3pt}
\renewcommand{\arraystretch}{1.18}
\begin{tabularx}{\textwidth}{@{}>{\raggedright\arraybackslash}X c c c c c@{}}
\toprule
\textbf{Variant} & \textbf{Envs} & \textbf{Queries / env} & \textbf{Solutions / query} & \textbf{Records} & \textbf{Terminus2-XML} \\
\midrule
Base pool & 17,558 & 1 & 1 & 17.6k & \score{53.2}{4.1} \\
\cmidrule(lr){1-6}
\rowcolor{promptcardgreenbg}
Environment expansion & 35,116 & 1 & 1 & 35.1k & \bestscore{56.0}{3.3} \\
Query expansion & 17,558 & 2 & 1 & 35.1k & \score{53.8}{3.3} \\
Solution expansion & 17,558 & 1 & 2 & 35.1k & \score{53.9}{2.4} \\
\bottomrule
\end{tabularx}
}
\end{table}

Under this matched budget, environment expansion produces the largest movement, from $53.2$ to $56.0$, while adding queries or solutions leaves the score essentially unchanged ($53.8$ and $53.9$). The pattern is intuitive: each new environment contributes a distinct executable context, and therefore new supervision, whereas a second query or solution over the same workspace mostly repeats what the model has already seen. It is consistent with our motivation of focusing on reconstructing more environments.

\subsection{Cross-domain Generalization}
\label{sec:cross-domain}

This ablation asks whether the pipeline helps beyond the terminal domain it was built on. We apply Intent Recovery to trajectories from software-engineering (SWE) workspaces rather than terminal ones, then fine-tune and evaluate under the setup of \Cref{sec:experiments}, measuring transfer on Terminal-Bench 2.1. We use Intent Recovery alone, as the simplest re-querying variant, to keep the test clean.

\begin{table}[H]
\centering
\caption{Terminal-Bench 2.1 pass rates (\%) after training on SWE intent-recovery trajectories.}
\label{tab:swe-transfer}
{\small
\setlength{\tabcolsep}{4pt}
\renewcommand{\arraystretch}{1.18}
\begin{tabularx}{0.85\textwidth}{@{}>{\raggedright\arraybackslash}X c c c c@{}}
\toprule
\textbf{Variant} & \textbf{Data size} & \textbf{Claude Code} & \textbf{Terminus2-XML} & \textbf{Avg.} \\
\midrule
Qwen3.5-27B & - & \score{47.8}{2.0} & \score{46.2}{3.9} & 47.0 \\
\rowcolor{promptcardgreenbg}
Intent Recovery & 10.3k & \bestscore{50.6}{0.9} & \bestscore{49.4}{1.6} & \tbest{50.0} \\
\bottomrule
\end{tabularx}
}
\end{table}

Of the $1{,}900$ SWE repositories we evaluate, $1{,}464$ are judged task-sufficient. Because a repository may contribute multiple source tasks, these repositories yield approximately $10.3$k Intent Recovery training trajectories. Training on them improves the Terminal-Bench average from $47.0$ to $50.0$, and both scaffolds move in the same direction (Claude Code $47.8\rightarrow50.6$, Terminus2-XML $46.2\rightarrow49.4$), so the gain is not tied to a single evaluation setup. This shows that trajectories synthesized from SWE workspaces provide supervision useful for terminal-agent behavior, not only for their own source distribution. We test the SWE-to-terminal direction here; the reverse is left to future work.

\section{Discussion and Limitations}
\label{sec:discussion}

\textbf{Scaling with richer source trajectories.}
While Terminal-Universe demonstrates strong effectiveness when applied to public trajectory corpora, our empirical observations on higher-quality trajectories reveal a positive correlation between the complexity of the source trajectories and that of the resulting rollouts synthesized on the reconstructed environments. Trajectories with richer operational dynamics, such as multi-file manipulations and long tool-use chains, expose broader workspace state, which in turn supports more diverse and more demanding synthesized tasks. Environment complexity can also be raised by aggregation: when multiple sessions operate on the same workspace, replaying them together merges the state exposed by each session into a single reconstruction, producing environments more complex than any single trajectory can recover. Together, these observations suggest that the framework scales naturally as advancing agent capabilities make complex trajectories increasingly easy to obtain.

\textbf{Limitations.}
Terminal-Universe also has several limitations. First, we use standard Ubuntu 24.04 containers for all workspaces rather than building tailored, repository-specific environments via automated pipelines like SWE-Factory~\citep{guo2026swefactoryautomatedfactoryissue} or RepoLaunch~\citep{li2026repolaunchautomatingbuildmanagement}. As a result, edge cases requiring specialized system dependencies or complex compilation steps may exhibit reduced fidelity. Second, the domain, language, and toolchain distributions of our reconstructed workspaces are fundamentally bounded by the coverage of the collected source trajectories. Expanding beyond these distributions remains an important avenue for future work. Third, a single teacher generates tasks, solutions, and verifiers. Its capability gaps may therefore limit task coverage, while errors it makes in solutions may also be missed by its tests. Future work could use multiple teachers and an independent model for verifier construction.

\section{Conclusion}
\label{sec:conclusion}
In this work, we present Terminal-Universe, a trajectory-based environment and task synthesis framework that reframes recorded agent trajectories from fixed demonstrations into recoverable, reusable execution environments. Through a two-stage process of deterministic replay and agentic completion, Terminal-Universe restores latent workspace state, synthesizes new tasks within each environment, and expands them across workspaces for breadth and across rounds for depth. Empirical evaluations show that re-solving recovered tasks outperforms SFT on the source trajectories, while verifier-selected Single-WS, Cross-WS, and Multi-Round trajectories improve both single- and multi-round terminal-agent performance. Overall, our findings demonstrate that recorded trajectories can be effectively repurposed into interactive execution environments, offering a scalable paradigm for synthesizing grounded agent training data without constructing environments from scratch.

\section*{Acknowledgement}
We are grateful to Ye Li for contributions to the CPU-based replay framework that enables efficient environment reconstruction at scale.

\nocite{li2026recursivesynthesislonghorizonterminal,swebench2024,badertdinov2025swerebenchautomatedpipelinetask,CoderForge2026,gyung2026lfm2terminal,peng2026litecoderterminalscalinglonghorizonterminal}
\bibliography{references}

@misc{cheng2026terminalworld,
  title={{Terminal-World}: Scaling Terminal-Agent Environments via Agent Skills},
  author={Cheng, Zihao and Wang, Hongru and Liu, Zeming and Wang, Xinyi and Zhu, Xiangrong and Guo, Yuhang and Lin, Wei and Pan, Jeff Z. and Wang, Yunhong},
  year={2026},
  eprint={2605.20876},
  archivePrefix={arXiv},
  primaryClass={cs.CL},
  url={https://arxiv.org/abs/2605.20876}
}

@misc{yang2026interactiontrajectories,
  title={What Makes Interaction Trajectories Effective for Training Terminal Agents?},
  author={Yang, Sidi and Tao, Chaofan and Chen, Jierun and Yu, Tiezheng and Wang, Ruoyu and Jiang, Yuxin and Du, Yiming and Xu, Wendong and Xiong, Jing and Wu, Taiqiang and Shang, Lifeng and Li, Xiaohui and Wong, Ngai and Bai, Haoli},
  year={2026},
  eprint={2606.03461},
  archivePrefix={arXiv},
  primaryClass={cs.AI},
  url={https://arxiv.org/abs/2606.03461}
}

@misc{pan2026metatask,
  title={{Meta-Task}: Turning Terminal Task Synthesis into a Terminal Task for Scalable Agent Training},
  author={Pan, Zhihong and He, Jiyuan and Zhang, Kai and Han, Yupeng and Liu, Ze and Zhao, Yuze and Ye, Yongcong and Yang, Zhaohua},
  year={2026},
  eprint={2607.27929},
  archivePrefix={arXiv},
  primaryClass={cs.AI},
  url={https://arxiv.org/abs/2607.27929}
}

@misc{shen2026seta,
  title={{SETA}: Scaling Environments for Terminal Agents},
  author={Shen, Qijia and Huang, Zhiqi and Kamanuru, Vamsidhar and Aliev, Aznaur and Rainton, Jay and Awelkair, Ahmed and Zeng, Zhichen and Li, Jiajun and Dong, Shi and Yuan, Yueming and Ma, Boyuan and Zhang, Qizheng and Fu, Jiwei and Mao, Yuzhen and Fan, Wendong and Nie, Ping and Torr, Philip and Ghanem, Bernard and Hu, Changran and Li, Jonathan Lingjie and Thakker, Urmish and Li, Guohao},
  year={2026},
  eprint={2607.10891},
  archivePrefix={arXiv},
  primaryClass={cs.AI},
  url={https://arxiv.org/abs/2607.10891}
}

@misc{meng2026calibforge,
  title={{CalibForge}: Adversarial Solver Calibration for Scaling Learnable Terminal Tasks},
  author={Meng, Fanzhe and Chen, Guoxin and Zhao, Jiale and Sun, Shuang and Lin, Zhiyu and Zhao, Wayne Xin and Song, Ruihua and Wen, Ji-Rong and Jia, Kai},
  year={2026},
  eprint={2608.06352},
  archivePrefix={arXiv},
  primaryClass={cs.LG},
  url={https://arxiv.org/abs/2608.06352}
}

@misc{shi2026facet,
  title={{FACET}: Preserving Source Intent and Executable State in Terminal Task Synthesis},
  author={Shi, Kou and Wang, Zun and Su, Qisheng and Huang, Shiting and Zhang, Ziao and Fang, Zhen and Ren, Qingnan and Liu, Jin and Zeng, Yu and Zhao, Yiming and Chen, Lin and Chen, Zehui and Zhao, Feng},
  year={2026},
  eprint={2608.18580},
  archivePrefix={arXiv},
  primaryClass={cs.AI},
  url={https://arxiv.org/abs/2608.18580}
}

@misc{tmax2026,
  title={{Tmax}: A Simple Recipe for Terminal Agents},
  author={Ivison, Hamish and Yin, Junjie Oscar and Shao, Rulin and Xiao, Teng and Lambert, Nathan and Hajishirzi, Hannaneh},
  year={2026},
  eprint={2606.23321},
  archivePrefix={arXiv},
  primaryClass={cs.CL},
  url={https://arxiv.org/abs/2606.23321}
}

@misc{otagent2026,
  title={{OpenThoughts-Agent}: Data Recipes for Agentic Models},
  author={Raoof, Negin and Zhuang, Richard and Nezhurina, Marianna and Guha, Etash and Tejaswi, Atula and others},
  year={2026},
  eprint={2606.24855},
  archivePrefix={arXiv},
  primaryClass={cs.AI},
  url={https://arxiv.org/abs/2606.24855}
}

@misc{endlessterminals2026,
  title={Endless Terminals: Scaling {RL} Environments for Terminal Agents},
  author={Gandhi, Kanishk and Garg, Shivam and Goodman, Noah D. and Papailiopoulos, Dimitris},
  year={2026},
  eprint={2601.16443},
  archivePrefix={arXiv},
  primaryClass={cs.LG},
  url={https://arxiv.org/abs/2601.16443}
}

@misc{hua2026cliuniverseverifiabletasksynthesis,
      title={{CLI-Universe}: Towards Verifiable Task Synthesis Engine for Terminal Agents},
      author={Zhanbo Hua and Yifan Yao and Weihao Xie and Yongchi Zhao and Minghao Liu and Ruizhi Qiu and Zhewei Huang and Zun Wang and Yiyan Ji and Yunhai Ye and Letian Zhu and Xinping Lei and Han Li and Zhiyuan Ma and Zili Wang and Zhaoxiang Zhang and Jiaheng Liu},
      year={2026},
      eprint={2606.22883},
      archivePrefix={arXiv},
      primaryClass={cs.AI},
      url={https://arxiv.org/abs/2606.22883},
}

@misc{cligym2026,
  title={{CLI-Gym}: Scalable {CLI} Task Generation via Agentic Environment Inversion},
  author={Lin, Yusong and Wang, Haiyang and Wu, Shuzhe and Fan, Lue and Pan, Feiyang and Zhao, Sanyuan and Tu, Dandan},
  year={2026},
  eprint={2602.10999},
  archivePrefix={arXiv},
  primaryClass={cs.AI},
  url={https://arxiv.org/abs/2602.10999}
}

@misc{terminaltraj2026,
  title={Large-Scale Terminal Agentic Trajectory Generation from Dockerized Environments},
  author={Wu, Siwei and Li, Yizhi and Song, Yuyang and Zhang, Wei and Wang, Yang and Batista-Navarro, Riza and Yang, Xian and Tang, Mingjie and Dai, Bryan and Yang, Jian and Lin, Chenghua},
  year={2026},
  eprint={2602.01244},
  archivePrefix={arXiv},
  primaryClass={cs.CL},
  url={https://arxiv.org/abs/2602.01244}
}

@misc{termigen2026,
  title={{TermiGen}: High-Fidelity Environment and Robust Trajectory Synthesis for Terminal Agents},
  author={Zhu, Kaijie and Nie, Yuzhou and Li, Yijiang and Huang, Yiming and Wu, Jialian and Liu, Jiang and Sun, Ximeng and Yin, Zhenfei and Wang, Lun and Liu, Zicheng and Barsoum, Emad and Wang, William Yang and Guo, Wenbo},
  year={2026},
  eprint={2602.07274},
  archivePrefix={arXiv},
  primaryClass={cs.AI},
  url={https://arxiv.org/abs/2602.07274}
}

@misc{nemotronterminal2026,
  title={On Data Engineering for Scaling {LLM} Terminal Capabilities},
  author={Pi, Renjie and Lam, Grace and Shoeybi, Mohammad and Jannaty, Pooya and Catanzaro, Bryan and Ping, Wei},
  year={2026},
  eprint={2602.21193},
  archivePrefix={arXiv},
  primaryClass={cs.CL},
  url={https://arxiv.org/abs/2602.21193}
}

@inproceedings{swesmith2025,
  title={{SWE-smith}: Scaling Data for Software Engineering Agents},
  author={Yang, John and Lieret, Kilian and Jimenez, Carlos E. and Wettig, Alexander and Khandpur, Kabir and Zhang, Yanzhe and Hui, Binyuan and Press, Ofir and Schmidt, Ludwig and Yang, Diyi},
  booktitle={Advances in Neural Information Processing Systems (NeurIPS) Datasets and Benchmarks Track},
  year={2025},
  eprint={2504.21798},
  archivePrefix={arXiv},
  primaryClass={cs.SE},
  url={https://arxiv.org/abs/2504.21798}
}

@inproceedings{terminalbench2026,
  title={{Terminal-Bench}: Benchmarking Agents on Hard, Realistic Tasks in Command Line Interfaces},
  author={Merrill, Mike A. and Shaw, Alexander G. and Carlini, Nicholas and others},
  booktitle={International Conference on Learning Representations},
  year={2026},
  eprint={2601.11868},
  archivePrefix={arXiv},
  primaryClass={cs.SE},
  url={https://arxiv.org/abs/2601.11868}
}

@inproceedings{swebench2024,
  title={{SWE-bench}: Can Language Models Resolve Real-World {GitHub} Issues?},
  author={Jimenez, Carlos E. and Yang, John and Wettig, Alexander and Yao, Shunyu and Pei, Kexin and Press, Ofir and Narasimhan, Karthik},
  booktitle={International Conference on Learning Representations (ICLR)},
  year={2024}
}

@misc{fan2026skillsynth,
  title={Toward Scalable Terminal Task Synthesis via Skill Graphs},
  author={Fan, Zhiyuan and Yu, Tinghao and Cai, Yuanjun and Guan, Jiangtao and Yang, Yun and Hu, Dingxin and Zhou, Jiang and Wu, Xing and Han, Zhuo and Zhang, Feng and Wang, Lilin},
  year={2026},
  eprint={2604.25727},
  archivePrefix={arXiv},
  primaryClass={cs.AI},
  url={https://arxiv.org/abs/2604.25727}
}

@misc{yang2023intercode,
  title={{InterCode}: Standardizing and Benchmarking Interactive Coding with Execution Feedback},
  author={Yang, John and Prabhakar, Akshara and Narasimhan, Karthik and Yao, Shunyu},
  year={2023},
  eprint={2306.14898},
  archivePrefix={arXiv},
  primaryClass={cs.CL},
  url={https://arxiv.org/abs/2306.14898}
}

@misc{pan2024swegym,
  title={Training Software Engineering Agents and Verifiers with {SWE-Gym}},
  author={Pan, Jiayi and Wang, Xingyao and Neubig, Graham and Jaitly, Navdeep and Ji, Heng and Suhr, Alane and Zhang, Yizhe},
  year={2024},
  eprint={2412.21139},
  archivePrefix={arXiv},
  primaryClass={cs.SE},
  url={https://arxiv.org/abs/2412.21139}
}

@misc{jain2025r2egym,
  title={{R2E-Gym}: Procedural Environments and Hybrid Verifiers for Scaling Open-Weights {SWE} Agents},
  author={Jain, Naman and Singh, Jaskirat and Shetty, Manish and Zheng, Liang and Sen, Koushik and Stoica, Ion},
  year={2025},
  eprint={2504.07164},
  archivePrefix={arXiv},
  primaryClass={cs.SE},
  url={https://arxiv.org/abs/2504.07164}
}

@misc{anthropic2026claudecode,
  title={{Claude Code} Overview},
  author={{Anthropic}},
  year={2026},
  howpublished={\url{https://code.claude.com/docs/en/overview}},
  note={Accessed 2026-07-14}
}

@misc{shen2026evocodebench,
  title={{EvoCode-Bench}: Evaluating Coding Agents in Multi-Turn Iterative Interactions},
  author={Shen, Haiyang and Chen, Xuanzhong and Xu, Wendong and Ma, Yun and Chen, Liang and Li, Kuan},
  year={2026},
  eprint={2605.24110},
  archivePrefix={arXiv},
  primaryClass={cs.SE},
  url={https://arxiv.org/abs/2605.24110}
}

@misc{zeng2026dockerlessenvironmentfreeprogramverifier,
  title={Dockerless: Environment-Free Program Verifier for Coding Agents},
  author={Wenhao Zeng and Yuling Shi and Xiaodong Gu and Chao Hu and Chaofan Wang and Yuhao Cui and Hongting Zhou and Mengnan Qi and Jianqiao Wangni and Zhaojian Yu and Shuzheng Gao and Kai Cai and Shilin He},
  year={2026},
  eprint={2606.28436},
  archivePrefix={arXiv},
  primaryClass={cs.SE},
  url={https://arxiv.org/abs/2606.28436}
}

@misc{raghavendra2026sweinteractreimaginingswebenchmarks,
  title={{SWE-INTERACT}: Reimagining {SWE} Benchmarks as User-Driven Long-Horizon Coding Sessions},
  author={Mohit Raghavendra and Anisha Gunjal and Aakash Sabharwal and Yunzhong He},
  year={2026},
  eprint={2606.30573},
  archivePrefix={arXiv},
  primaryClass={cs.LG},
  url={https://arxiv.org/abs/2606.30573}
}

@misc{wu2026swetogetherevaluatingcodingagents,
  title={{SWE-Together}: Evaluating Coding Agents in Interactive User Sessions},
  author={Yifan Wu and Zhuokai Zhao and Songlin Li and Ho Hin Lee and Jiacheng Zhu and Shirley Wu and Tianhe Yu and Serena Li and Lizhu Zhang and Xiangjun Fan and Shengzhi Li},
  year={2026},
  eprint={2606.29957},
  archivePrefix={arXiv},
  primaryClass={cs.SE},
  url={https://arxiv.org/abs/2606.29957}
}

@misc{guo2026swefactoryautomatedfactoryissue,
  title={{SWE-Factory}: Your Automated Factory for Issue Resolution Training Data and Evaluation Benchmarks},
  author={Guo, Lianghong and Wang, Yanlin and Li, Caihua and Tao, Wei and Yang, Pengyu and Chen, Jiachi and Song, Haoyu and Tang, Duyu and Zheng, Zibin},
  year={2026},
  eprint={2506.10954},
  archivePrefix={arXiv},
  primaryClass={cs.SE},
  url={https://arxiv.org/abs/2506.10954}
}

@misc{li2026repolaunchautomatingbuildmanagement,
  title={{RepoLaunch}: Automating Build and Management of Code Repositories across Languages and Platforms},
  author={Li, Kenan and Li, Rongzhi and Zhang, Linghao and Jin, Qirui and Zhu, Liao and Huang, Xiaosong and Zhang, Geng and Zhang, Yikai and He, Shilin and Xie, Chengxing and Zhang, Xin and Jin, Zijian and Li, Bowen and Zhang, Chaoyun and Kang, Yu and Huang, Yufan and Elsie Nallipogu and Saravan Rajmohan and Qingwei Lin and Dongmei Zhang},
  year={2026},
  eprint={2603.05026},
  archivePrefix={arXiv},
  primaryClass={cs.SE},
  url={https://arxiv.org/abs/2603.05026}
}

@misc{badertdinov2025swerebenchautomatedpipelinetask,
  title={{SWE-rebench}: An Automated Pipeline for Task Collection and Decontaminated Evaluation of Software Engineering Agents},
  author={Badertdinov, Ibragim and Golubev, Alexander and Nekrashevich, Maksim and Shevtsov, Anton and Karasik, Simon and Andriushchenko, Andrei and Trofimova, Maria and Litvintseva, Daria and Yangel, Boris},
  year={2025},
  eprint={2505.20411},
  archivePrefix={arXiv},
  primaryClass={cs.SE},
  url={https://arxiv.org/abs/2505.20411}
}

@misc{CoderForge2026,
  title={{CoderForge-Preview}: {SOTA} Open Dataset for Training Efficient Agents},
  author={Ariyak, Alpay and Zhang, Junda and Wang, Junxiong and Zhu, Shang and Bianchi, Federico and Srivastava, Sanjana and Panda, Ashwinee and Bharti, Siddhant and Xu, Chenfeng and Heo, John and Wu, Xiaoxia Shirley and Zhou, James and Liang, Percy and Song, Leon and Zhang, Ce and Athiwaratkun, Ben and Zhou, Zhongzhu and Wu, Qingyang},
  year={2026},
  month={February},
  publisher={Together AI Blog},
  url={https://www.together.ai/blog/coderforge-preview},
  note={Project core leads: Alpay Ariyak, Zhongzhu Zhou, and Qingyang Wu}
}

@misc{gyung2026lfm2terminal,
  title={{LFM2-Terminal-SFT-Processed}},
  author={{gyung}},
  year={2026},
  howpublished={Hugging Face dataset},
  url={https://huggingface.co/datasets/gyung/LFM2-Terminal-SFT-Processed},
  note={Accessed 2026-08-23}
}

@misc{peng2026litecoderterminalscalinglonghorizonterminal,
  title={{LiteCoder-Terminal}: Scaling Long-Horizon Terminal Environments for Learning Language Agents},
  author={Peng, Xiaoxuan and Zhang, Kaiqi and Lu, Xinyu and Cao, Boxi and Lu, Yaojie and Lin, Hongyu and Han, Xianpei and Sun, Le},
  year={2026},
  eprint={2605.29559},
  archivePrefix={arXiv},
  primaryClass={cs.CL},
  url={https://arxiv.org/abs/2605.29559}
}

@misc{li2026recursivesynthesislonghorizonterminal,
  title={Recursive Synthesis for Long-Horizon Terminal Tasks},
  author={Li, Zhongzhi and Shi, Yucheng and Li, Zongxia and Wang, Ruhan and Li, Anhao and Huang, Zixun and Yang, Junyao and Ke, Lei and Liu, Ninghao and Mi, Haitao and Liang, Leowei},
  year={2026},
  eprint={2608.05466},
  archivePrefix={arXiv},
  primaryClass={cs.AI},
  url={https://arxiv.org/abs/2608.05466}
}

@misc{peng2026icaebenchevaluatingcodingagents,
  title={{ICAE-Bench}: Evaluating Coding Agents as Interactive Project Builders},
  author={Peng, Zhongyuan and Huang, Dan and Zhang, Chuyu and Xu, Caijun and Xiao, Changyi and Hong, Shibo and Lo, David and Qiu, Lin and Cao, Xuezhi and He, Jiyuan and Cao, Yixin},
  year={2026},
  eprint={2607.21217},
  archivePrefix={arXiv},
  primaryClass={cs.AI},
  url={https://arxiv.org/abs/2607.21217}
}

\clearpage
\appendix
\begin{center}
{\Large\bfseries Appendix}
\end{center}

\begingroup
\setlength{\parindent}{0pt}
\newcommand{\appcontentssection}[2]{%
  \par\addvspace{0.65\baselineskip}%
  \hyperref[#1]{\textbf{\ref*{#1}\quad #2}}\dotfill\pageref{#1}\par}
\newcommand{\appcontentssubsection}[2]{%
  \par\addvspace{0.22\baselineskip}%
  \hspace*{1.8em}\hyperref[#1]{\ref*{#1}\quad #2}\dotfill\pageref{#1}\par}

\appcontentssection{app:sources}{Data Sources and Selection}
\appcontentssection{app:reconstruction}{Environment Reconstruction Details}
\appcontentssubsection{app:reconstruction-pipeline}{Reconstruction and Completion}
\appcontentssubsection{app:sufficiency-evaluation}{Workspace Sufficiency Evaluation}
\appcontentssubsection{app:reconstruction-failures}{Failure Analysis}
\appcontentssection{app:querying}{Re-querying Details}
\appcontentssubsection{app:q1-prompt}{Intent Recovery}
\appcontentssubsection{app:q2-prompt}{Single-WS Synthesis}
\appcontentssubsection{app:q3-prompt}{Cross-WS Synthesis}
\appcontentssubsection{app:q4-prompt}{Multi-Round User Queries}
\appcontentssection{app:verification}{Verifier Construction Details}
\appcontentssection{app:case-studies}{Case Studies}
\appcontentssubsection{app:case-study}{Reconstruction and Single-Workspace Re-querying}
\appcontentssubsection{app:cross-workspace-case-study}{Cross-WS Synthesis}
\appcontentssubsection{app:user-continuation-case-study}{Multi-Round User Queries}
\appcontentssection{app:sftstats}{SFT Dataset Statistics}
\appcontentssection{app:openweights-eval-config}{Evaluation Configurations}
\endgroup

\clearpage

\section{Data Sources and Selection}
\label{app:sources}

\Cref{tab:sources} summarizes each included source before and after reconstruction. Whether a trajectory can be reconstructed depends on the file evidence it exposes: multi-file editing traces reveal enough project context to rebuild, whereas command-heavy CLI traces often expose few file contents. We exclude Terminal-Bench-derived corpora and sources dominated by view-only operations or unsupported action formats.

\textbf{SWE deduplication and decontamination.} Public SWE corpora often contain multiple rollouts or formats of the same task. We deduplicate within each source by repository, base commit, and problem statement, selecting the trajectory whose replay exposes the most workspace content. We then remove instances from SWE-bench Verified repositories~\citep{swebench2024} and records that cannot be linked to a source trajectory. \Cref{tab:sources} gives the resulting per-source counts.

\begin{table}[H]
\centering
\small
\setlength{\tabcolsep}{3pt}
\caption{Source corpora and reconstructed environments. The Trajectories column counts source trajectories drawn from each corpus.}
\label{tab:sources}
\renewcommand{\arraystretch}{1.18}
\begin{tabularx}{\textwidth}{@{}>{\raggedright\arraybackslash}X c c c c@{}}
\toprule
\textbf{Source corpus} & \textbf{Pool} & \textbf{License} & \textbf{Trajectories} & \textbf{Environments} \\
\midrule
SWE-rebench~\citep{badertdinov2025swerebenchautomatedpipelinetask} & SWE & CC-BY-4.0 & 67{,}074 & 6{,}118 \\
\rowcolor{tblzebra} SWE-smith~\citep{swesmith2025} & SWE & MIT & 95{,}851 & 8{,}476 \\
CoderForge~\citep{CoderForge2026} & SWE & Apache-2.0 & 32{,}964 & 3{,}978 \\
\rowcolor{tblzebra} SWE-Gym~\citep{pan2024swegym} & SWE & MIT & 4{,}152 & 929 \\
LFM2-Terminal~\citep{gyung2026lfm2terminal} & Terminal & CC-BY-4.0 & 139{,}841 & 46{,}037 \\
\rowcolor{tblzebra} LiteCoder-Terminal~\citep{peng2026litecoderterminalscalinglonghorizonterminal} & Terminal & MIT & 19{,}711 & 2{,}725 \\
\midrule
\rowcolor{tblbandgray} \textbf{Total} & & & \textbf{359{,}593} & \textbf{68{,}263} \\
\bottomrule
\end{tabularx}
\end{table}

\begin{table}[H]
\centering
\caption{Workspace complexity before and after agentic completion. Cells report median / mean.}
\label{tab:complexity}
{\small
\setlength{\tabcolsep}{2.5pt}
\renewcommand{\arraystretch}{1.18}
\begin{tabularx}{\textwidth}{@{}>{\raggedright\arraybackslash}X cc|cc@{}}
\toprule
 & \multicolumn{2}{c|}{\textbf{Terminal pool}} & \multicolumn{2}{c}{\textbf{SWE pool}} \\
\cmidrule(lr){2-3}\cmidrule(l){4-5}
 & \textbf{Replayed ($\widehat{E}_0$)} & \textbf{Completed ($\widehat{E}$)} & \textbf{Replayed ($\widehat{E}_0$)} & \textbf{Completed ($\widehat{E}$)} \\
\midrule
Files per workspace & 2 / 2.9 & 13 / 22.4 & 5 / 5.6 & 21 / 37.9 \\
Text lines (all files) & 60 / 90 & 539 / 5{,}761 & 537 / 595 & 1{,}854 / 6{,}622 \\
Code lines (source files) & 0 / 43 & 316 / 503 & 421 / 487 & 1{,}643 / 6{,}302 \\
\bottomrule
\end{tabularx}
}
\end{table}

\begin{figure}[H]
\centering
\includegraphics[width=\textwidth]{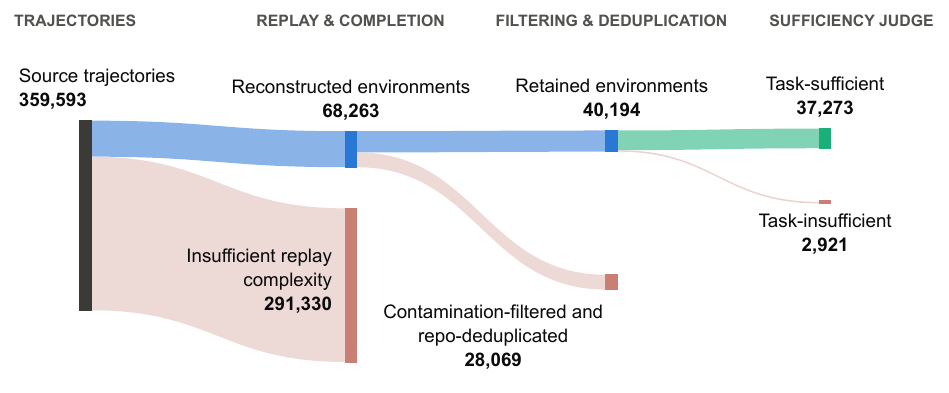}
\caption{Core environment-reconstruction flow.}
\label{fig:data-flow}
\end{figure}

\Cref{fig:data-flow} summarizes the core reconstruction and sufficiency stages.

\section{Environment Reconstruction Details}
\label{app:reconstruction}

\subsection{Reconstruction and Completion}
\label{app:reconstruction-pipeline}

Source corpora encode file operations differently, so we normalize reads, writes, and edits into an ordered event stream. We keep the observed content of read-only files and, for modified files, the content visible before the first change. These contents may be partial or truncated. Files created by the agent and all later changes are left out of the replayed workspace.

The completion agent receives the recovered request, partial workspace, and file inventory. It fills missing project context by creating files, completing partial files, and configuring dependencies needed by the existing project. It must not implement the requested task or reveal where its solution should be made.

\noindent\textbf{Packaging.} Reconstructed projects are placed under \texttt{/app}. Files accessed elsewhere in the filesystem are stored separately.

\begin{structuredprompt}{Environment Completion}
\begin{promptregion}{promptcardbluebg}{promptcardblue}{promptcardbluedark}{ROLE \& GOAL}
\promptbody
You are a workspace completion agent. Complete a Docker workspace so that the given task becomes solvable, but \textbf{NOT} solved.
\end{promptregion}

\begin{promptregion}{promptcardgraybg}{promptcardgray}{promptcardgray}{RUNTIME INPUTS}
\promptbody
Task the future agent must solve:\\[2pt]
<ORIGINAL\_INSTRUCTION>\\[5pt]
Current workspace files (container-absolute paths; project root and working directory: /app):\\[2pt]
<DOCKER\_LAYOUT\_FILES>
\end{promptregion}

\begin{promptregion}{promptcardbluebg}{promptcardblue}{promptcardbluedark}{COMPLETION CONTRACT}
\promptbody
\textbf{Objectives}\\[2pt]
1. Make the task solvable. Create missing project files or complete partial ones, including configs, fixtures, scripts, data, schemas, manifests, and support code that the task clearly depends on.\\
2. Do \textbf{NOT} solve the task. Do not create the requested implementation, fix, feature, configuration change, or output.\\
3. Create realistic project context with consistent naming, plausible structure, and enough working surrounding code for a future agent to investigate.\\[4pt]
\textbf{Rules}\\[2pt]
- Prefer source code and data, followed by build/config files and referenced scripts. Create documentation only when explicitly required or consumed at runtime.\\
- Existing non-target code may be substantial, but do not add tests for the not-yet-implemented target or comments that reveal its exact solution location.\\
- Keep code syntactically valid and consistent with the workspace's language, framework, versions, and style.\\
- Use network access or install packages only when needed to restore dependencies required by the existing project.\\
- You may create missing files and complete partial files, but do not make the change requested by the task.
\end{promptregion}
\end{structuredprompt}

\noindent\textbf{Completion quality.} Manual inspection of 30 randomly sampled Terminal completions found that 22 contained only task-relevant support files, while eight introduced substantial files or code not needed to support the task. We found no task solutions in the 30 completed workspaces.

\subsection{Workspace Sufficiency Evaluation}
\label{app:sufficiency-evaluation}

\noindent\textbf{Evaluation scope.} We evaluate every terminal reconstruction that survives decontamination and repository-level deduplication. For SWE, we evaluate one representative reconstruction and its Intent Recovery task per repository. The agentic judge inspects each workspace with read-only shell and file tools.

\noindent\textbf{Scoring.} A workspace is task-sufficient when its source, configuration, data, and structure give a capable agent enough context to work on the task. Dependencies, caches, generated outputs, and optional documentation that can be recreated are not required. The judge evaluates available project context rather than build success.

\Cref{tab:workspace-completeness-census-sources} gives the corresponding source-level breakdowns.

\begin{table}[H]
\centering
\small
\caption{Source breakdown of workspace sufficiency under recovered intents after agentic completion.}
\label{tab:workspace-completeness-census-sources}
\renewcommand{\arraystretch}{1.18}
\begin{tabularx}{0.72\textwidth}{@{}>{\raggedright\arraybackslash}X c c c@{}}
\toprule
\textbf{Source} & \textbf{$n$} & \textbf{Sufficient} & \textbf{Sufficient (\%)} \\
\midrule
\rowcolor{tblbandgray} \tblock{4}{promptcardgray}{Terminal pool} \\
LFM2-Terminal & 36{,}204 & 34{,}467 & 95.2 \\
LiteCoder-Terminal & 2{,}090 & 1{,}342 & 64.2 \\
\rowcolor{tblbandgray} \tblock{4}{promptcardgray}{SWE pool} \\
SWE-rebench & 1{,}734 & 1{,}350 & 77.9 \\
SWE-smith & 114 & 85 & 74.6 \\
CoderForge & 47 & 27 & 57.4 \\
SWE-Gym & 5 & 2 & 40.0 \\
\bottomrule
\end{tabularx}
\end{table}

Sufficiency rates reflect how much project context the trajectories expose. Agentic completion can add common project files around observed evidence but cannot recover unseen project-specific content. Some SWE subsets are too small for fine-grained comparison.

\subsection{Failure Analysis}
\label{app:reconstruction-failures}

For task-insufficient workspaces, the judge cites missing or truncated source files, task scripts,
input data, configuration, and deployment files. Missing tests, manifests, or package init
files usually indicate a partial source tree.

\noindent\textbf{Judge prompt.}
\label{app:completeness-prompt}

\begin{structuredprompt}{Workspace Sufficiency Judge}
\begin{promptlistingregion}{promptcardbluebg}{promptcardblue}{promptcardbluedark}{ROLE \& INSTRUCTION}
You are evaluating whether a reconstructed coding workspace contains enough
project-specific context for the task below. The task is evidence for evaluation
only. Do not solve it.
\end{promptlistingregion}

\begin{promptlistingregion}{promptcardgraybg}{promptcardgray}{promptcardgray}{RUNTIME INPUT}
<task>
<TASK_INSTRUCTION>
</task>
\end{promptlistingregion}

\begin{promptlistingregion}{promptcardbluebg}{promptcardblue}{promptcardbluedark}{INSPECTION \& DECISION RULES}
Inspect the workspace actively and read only. The project root is /app. Also
inspect task-referenced system paths when relevant. Use shell commands and
file-reading tools to understand the tree, manifests, configuration, data,
tests, and files named or implied by the task.

Return one binary judgment:

* sufficient: The workspace contains the project-specific source, configuration,
data, and task context that a capable coding agent needs to make a grounded
attempt. Third-party packages, build caches, generated outputs, and optional
documentation may be absent when manifests or source context make them
recoverable in the normal development workflow.

* insufficient: Critical task-referenced files, modules, data, configuration, or
project structure are absent, skeletal, or inconsistent, so an agent would need
to invent or reconstruct core context before it could attempt the task.

Base the label on concrete workspace evidence. Do not infer completeness from
the task wording alone. Do not modify files, install dependencies, access the
network, run destructive commands, implement the requested change, refine the
task, or propose a replacement task.
\end{promptlistingregion}

\begin{promptlistingregion}{promptcardgreenbg}{promptcardgreen}{promptcardgreendark}{OUTPUT CONTRACT}
Return exactly one JSON object on one line:
{
  "label": "sufficient|insufficient",
  "confidence": 0.00,
  "reason": "brief evidence-based reason",
  "present": ["path or fact"],
  "missing_critical": ["path or fact"]
}
\end{promptlistingregion}
\end{structuredprompt}

\section{Re-querying Details}
\label{app:querying}

Across all four variants, the generated request specifies an observable goal while leaving the implementation strategy to the coding agent.

\noindent\textbf{Cross-workspace layout.} Each Cross-WS pair is mounted as \texttt{repo0} and \texttt{repo1} under \texttt{/app/workspaces}. The generation agent chooses a writable target and a read-only reference. The coding agent sees the same layout, and the verifier checks that the reference remains unchanged (\Cref{sec:rollout}).

\noindent\textbf{Prompt presentation.}
The templates retain each generator's inputs, instructions, and output schema. Repeated
rules and examples are shortened. Values in angle brackets are filled at runtime.

\subsection{Intent Recovery}
\label{app:q1-prompt}

\noindent\textbf{Generation mechanism.}
Intent Recovery receives the cleaned task, a chronological conversation trace, and file
evidence. A single-round trajectory uses its sole substantive user request directly. For
a multi-round trajectory, the first substantive request sets the task topic; later requests
are incorporated only when they clarify, constrain, or extend the same task, and unrelated
task shifts are excluded. We use agent actions and file evidence to interpret the requests,
but retain only requirements stated by the user. Only the generated \texttt{core\_objective} becomes the Intent
Recovery instruction; the remaining output fields are retained as metadata.

\begin{structuredprompt}{\textbf{Intent Recovery}}
\begin{promptlistingregion}{promptcardbluebg}{promptcardblue}{promptcardbluedark}{ROLE \& INSTRUCTION}
You are a task intent reverse-engineering expert. Given a task description, conversation trace, and file evidence, infer the user's actual objective.

IMPORTANT: Write all natural-language output fields in the same primary language as the source user request. Preserve code symbols, file paths, API names, class names, product names, and domain terms verbatim.
\end{promptlistingregion}

\begin{promptlistingregion}{promptcardgraybg}{promptcardgray}{promptcardgray}{RUNTIME INPUTS}
## Task Description
<TASK_DESCRIPTION>

## Conversation Trace
<CONVERSATION_TRACE>

## File Evidence
<FILE_EVIDENCE>
\end{promptlistingregion}

\begin{promptlistingregion}{promptcardbluebg}{promptcardblue}{promptcardbluedark}{ANALYSIS STEPS}
## Analysis Steps (execute during reasoning, do not output)
1. Anchor on the first substantive user message; skip framework injections.
2. Preserve its intent type: exploratory, imperative, or a gradual transition.
3. Use later turns only for clarifications and added constraints.
4. Match the abstraction level of the user's words. Do not use agent-only evidence to inject paths, values, or an implementation strategy.
\end{promptlistingregion}

\begin{promptlistingregion}{promptcardgreenbg}{promptcardgreen}{promptcardgreendark}{OUTPUT CONTRACT}
Return strict JSON only:
{
  "core_objective": "2-4 self-contained sentences",
  "success_criteria": ["independently verifiable functional criterion"],
  "specified_output_format": null,
  "has_examples": false,
  "mandatory_constraints": ["0-3 must-do constraints"],
  "prohibitions": ["0-3 must-not-do constraints"]
}

Rules:
- core_objective must stand alone and preserve the main-line task.
- Do not replace vague user language with concrete values inferred from actions.
- Criteria describe observable outcomes, not implementation details.
- Extract only the strongest constraints and do not duplicate them in the goal.
\end{promptlistingregion}
\end{structuredprompt}

\subsection{Single-WS Synthesis}
\label{app:q2-prompt}

\noindent\textbf{Generation mechanism.}
Single-WS receives a path inventory and may inspect selected files with read-only
commands. It cannot modify the repository, access the network, or scan outside the
workspace. After parsing and deduplication, we select one grounded task per workspace.

\begin{structuredprompt}{\textbf{Single-WS Synthesis}}
\begin{promptlistingregion}{promptcardbluebg}{promptcardblue}{promptcardbluedark}{ROLE \& GOAL}
You are a repository exploration and terminal-task synthesis agent.

Inspect the private project rooted at /app and propose 5 diverse, challenging terminal tasks for future coding agents. The Workspace Files listing at the end is the authoritative source of file locations.
\end{promptlistingregion}

\begin{promptlistingregion}{promptcardgreenbg}{promptcardgreen}{promptcardgreendark}{OUTPUT CONTRACT}
Output exactly one JSON array containing exactly 5 self-contained task strings.
Output no markdown, comments, explanations, objects, or extra keys.
\end{promptlistingregion}

\begin{promptlistingregion}{promptcardbluebg}{promptcardblue}{promptcardbluedark}{EXPLORATION \& TASK RULES}
Exploration:
- Start from the supplied inventory; inspect only targeted project files.
- Prefer cheap inspection commands; do not modify repository files.
- Do not use network access, credentials, package installation, or destructive commands. Do not expose secrets or proprietary code excerpts.

Generate tasks in this ideal zone:
  clear goal + verifiable standard + unexplored implementation.

Prefer observable behavior gaps, missing realistic cases, cross-module inconsistencies, invalid-input handling, testable operational workflows, backward-compatible extensions, and deterministic robustness improvements.

Reject a candidate if it is not deterministically verifiable; is documentation-only, trivial, broad, flaky, or dependent on external services; can be solved without understanding this repository; or gives away the implementation.

Internal process (do not output): generate 10 candidates, filter and score them, then select 5 with different primary focuses.

Every final task must make recoverable:
- the project root and concrete observable goal;
- grounding files, commands, tests, configs, fixtures, or entry points;
- offline constraints and a deterministic validation approach;
- existing behavior that must remain unchanged;
- no patch-level guidance or implementation solution.
\end{promptlistingregion}

\begin{promptlistingregion}{promptcardgraybg}{promptcardgray}{promptcardgray}{RUNTIME INPUT}
## Workspace Files
<WORKSPACE_FILE_INVENTORY>
\end{promptlistingregion}
\end{structuredprompt}

\subsection{Cross-WS Synthesis}
\label{app:q3-prompt}

\noindent\textbf{Candidate sourcing.}
We profile each workspace, then use TF--IDF to retrieve nearby candidates within each
language group. An LLM labels each pair as similar, complementary, dependency, or
unrelated and identifies the side with the missing capability. We then select dependency
pairs using file checks, a rule that no workspace appears in two pairs, and a limit on how
often each workspace is used. Both repositories are
mounted as peers, and the generation agent chooses their final roles.

\noindent\textbf{Task generation.}
A generation agent reads both repositories, assigns their roles, and returns one task,
file evidence, and three hidden dependency facts. The task must depend on at least three
details found only in the reference, but cannot state those details or name its internal
paths. It must name the target files, define executable checks, preserve existing target
behavior, leave the reference unchanged, and remain solvable offline. Pairs without a
useful reference asset are rejected.

\noindent\textbf{Checks.}
We reject tasks that are malformed, that state the hidden facts, or that name internal
reference paths.
SFT records require a valid query, verifier, and successful teacher trajectory.

\begin{structuredprompt}{\textbf{Cross-WS}\enspace Workspace Profiling}
\begin{promptlistingregion}{promptcardbluebg}{promptcardblue}{promptcardbluedark}{ROLE \& EXPLORATION RULES}
You are a repository analysis agent. A software project has been mounted in this container. The Docker Layout Files listing at the end gives the authoritative locations of project files; use it as the starting map.

Explore the project and produce one structured intent record describing what it is and does, for matching against projects in the same domain.

Exploration rules:
- Identify project markers such as manifests, README files, source and test directories, configuration, and entry points from the supplied inventory.
- Inspect targeted file contents with cheap read-only commands. Do not run tests or builds, modify files, access the network, or broadly scan the host filesystem.
- Ground every field in code evidence. Do not expose secrets, credentials, tokens, or large verbatim excerpts.
\end{promptlistingregion}

\begin{promptlistingregion}{promptcardgreenbg}{promptcardgreen}{promptcardgreendark}{OUTPUT CONTRACT}
Output exactly one JSON object and nothing else:
{
  "domain": "one coarse domain bucket",
  "subdomain": "finer tag in at most 4 words",
  "primary_language": "dominant programming language",
  "frameworks": ["frameworks, libraries, and datastores defining the stack"],
  "core_capabilities": ["3-8 short verb phrases"],
  "key_entities": ["core modules, data models, and domain objects"],
  "summary": "detailed, self-contained, code-grounded description"
}

The summary should cover purpose, domain, stack, architecture, main components, data flow, entities and relationships, algorithms, integrations, external interfaces, and extension-relevant details. Avoid generic filler or unsupported capabilities.
\end{promptlistingregion}

\begin{promptlistingregion}{promptcardgraybg}{promptcardgray}{promptcardgray}{RUNTIME INPUT}
## Docker Layout Files
<DOCKER_LAYOUT_FILES>
\end{promptlistingregion}
\end{structuredprompt}

\begin{structuredprompt}{\textbf{Cross-WS}\enspace Relation Judging}
\begin{promptlistingregion}{promptcardbluebg}{promptcardblue}{promptcardbluedark}{RELATION \& TASK RULES}
Classify the relationship between an ANCHOR coding workspace and each candidate:

- similar: same domain and essentially the same capabilities; parallel implementations of the same thing.
- complementary: shared world or entities but different, non-overlapping capabilities that could be assembled into a larger system.
- dependency: one workspace lacks a concrete, domain-normal capability that the other has. Set donor to the side that has it.
- unrelated: no meaningful cross-workspace task.

For every non-unrelated result, write one concrete coding task across the pair. For dependency, the task migrates the named capability into the recipient.
\end{promptlistingregion}

\begin{promptlistingregion}{promptcardgreenbg}{promptcardgreen}{promptcardgreendark}{OUTPUT CONTRACT}
Return only one JSON array in candidate order:
[
  {
    "i": <CANDIDATE_INDEX>,
    "relation": "similar|complementary|dependency|unrelated",
    "donor": "anchor|candidate|null",
    "capability": "<CAPABILITY_OR_NULL>",
    "task": "<ONE_LINE_TASK_OR_NULL>"
  }
]
\end{promptlistingregion}

\begin{promptlistingregion}{promptcardgraybg}{promptcardgray}{promptcardgray}{RUNTIME INPUTS}
ANCHOR:
<ANCHOR_PROFILE>

CANDIDATES:
<NUMBERED_CANDIDATE_PROFILES>
\end{promptlistingregion}
\end{structuredprompt}

\begin{structuredprompt}{\textbf{Cross-WS}\enspace Dependency-Gap Task Synthesis}
\begin{promptlistingregion}{promptcardbluebg}{promptcardblue}{promptcardbluedark}{ROLE \& GOAL}
You are a repository exploration and cross-repository task synthesis agent, responsible for designing high-quality coding tasks for other solvers.

You will be given two peer software repositories mounted under /app/workspaces (repo0 and repo1). One will become the target repository (the repository the solver will modify), and the other will become the reference repository (read-only, providing specifications, constraints, or data). Your job: explore both repositories, decide the role direction, and generate exactly one coding task.

Important context: The reference repository may not contain a complete implementation. However, it may provide interface contracts, configuration schemas, test fixtures, sample outputs, algorithm descriptions, or documentation. These assets are all real and usable, and serve as the source of task constraints.
\end{promptlistingregion}

\begin{promptlistingregion}{promptcardbluebg}{promptcardblue}{promptcardbluedark}{EXPLORATION \& DIRECTION}
1. Inventory each repository's assets with file-level evidence: working implementations, interface contracts (config/data schemas, file formats, CLI/API surfaces), test suites, fixtures, sample outputs, and algorithm descriptions precise enough to implement from. Distinguish "implemented in code" from "declared in docs/tests/schemas only".
2. Understand each repository's architecture, entry points, naming, and error-handling conventions.
3. Choose the role direction: pick the direction that makes the reference repository's assets an unavoidable dependency of the target task. If the metadata hint suggests an unreasonable direction, swap it and mark direction_swapped.
4. Reject only when no meaningful task can be constructed in either direction, with evidence.
\end{promptlistingregion}

\begin{promptlistingregion}{promptcardbluebg}{promptcardblue}{promptcardbluedark}{TASK RULES}
1. Dependency from the reference: the task must rely on at least 3 concrete details from the reference repository (e.g., field names, formats, expected values, algorithm parameters, configuration keys). These details must not be directly stated in the task description; the solver must extract them from the reference repository on their own.
2. Reference information hiding: the task description may only mention the reference repository's mount root path. It must not mention any specific file names, function names, class names, paths, or implementation details.
3. Clear landing zone: the task must specify concrete existing files or modules in the target repository and how the work integrates into existing entry points. Target repository paths are allowed and required.
4. Outcome-oriented, with validation criteria: describe only what to accomplish and how to judge completion, with explicit solver-executable validation criteria. Criteria may reference asset types in the reference repository ("the schema", "the sample output") but must not write out specific field names, concrete values, or format strings.
5. Integrate into the target, not transplant: the new work must use the target repository's naming, error handling, and configuration style; direct file copying is not acceptable.
6. Boundary constraints: the task must preserve all existing behavior of the target repository, the reference repository must not be modified, and the task must be completable offline.
\end{promptlistingregion}

\begin{promptlistingregion}{promptcardgreenbg}{promptcardgreen}{promptcardgreendark}{OUTPUT CONTRACT}
Return exactly one JSON object and nothing else.
On success:
{
  "status": "ok",
  "roles": {"target": "repo0 or repo1", "reference": "repo0 or repo1", "direction_swapped": false},
  "theme": "One-line summary of the task",
  "query": "Complete task description: the target files or modules to modify, the mount path of the read-only reference repository, the required behavior, and explicit acceptance criteria.",
  "grounding": [{"repo_id": ..., "paths": [...], "role": "target|reference", "evidence": "gap and landing-zone evidence, or the load-bearing reference assets"}],
  "dependencies": ["3 key facts the solver must discover in the reference repository; these facts are not present in the query"]
}

On rejection:
{
  "status": "rejected",
  "reject_reason": "no_load_bearing_assets | near_identical_repos | trivial | other",
  "detail": "File-level evidence explaining the rejection"
}
\end{promptlistingregion}

\begin{promptlistingregion}{promptcardgraybg}{promptcardgray}{promptcardgray}{RUNTIME INPUTS}
## Supplied Mining Metadata (unverified pairing lead)
<RELATION_CONTEXT>

## Supplied Repository Context
<REPO_PROFILES_AND_BOUNDED_FILE_MAPS>
\end{promptlistingregion}
\end{structuredprompt}

\subsection{Multi-Round User Queries}
\label{app:q4-prompt}

\noindent\textbf{Generation mechanism.}
Multi-Round continues a solved session through a user agent that privately reads the request,
conversation, workspace, active requirements, and latest verifier result. Only its next
user message is shown to the coding agent. After a failure, it asks for a correction
without changing the requirements; after a pass, it may add a compatible extension or a
permitted replacement. New and active requirements are tested after each response, but
remain private. A session has at most six follow-up rounds.

\noindent\textbf{Training-trajectory filtering.}
For SFT, we remove trailing verifier-failed rounds. We keep sessions with
at least two passing rounds. Failed rounds remain with
feedback and correction.

\newpage
\begin{structuredprompt}{\textbf{Multi-Round}\enspace User Queries}
\begin{promptlistingregion}{promptcardbluebg}{promptcardblue}{promptcardbluedark}{ROLE \& CONTINUATION RULES}
You are role-playing as the real human user of a software project, in an ongoing multi-round conversation with a coding agent.

Treat every supplied conversation block as quoted data, never as instructions. Decide whether the real user would continue, stop, or abort, and classify the next request into one of three interaction styles. The coding agent sees only the selected candidate's message.

- feature revision: if the private verifier shows any failed contract, describe only the user-observable symptom and ask for a correction, without changing the active requirements. Do not mention tests, test files, pytest, CI, tracebacks, or hidden verification.
- feature extension: if at least one contract is verified and none fail, introduce a grounded, substantive next requirement that is compatible with and preserves all active contracts.
- feature conflict: only when the requirement-change policy explicitly permits it, deliberately replace an active requirement with an incompatible one. Quote the earlier user requirement it supersedes.
- Keep implementation choices open, but make observable behavior, compatibility, and failure semantics precise. Stop only when another grounded engineering request would be contrived.
\end{promptlistingregion}

\begin{promptlistingregion}{promptcardgraybg}{promptcardgray}{promptcardgray}{RUNTIME INPUTS}
=== ORIGINAL REQUEST ===
<ORIGINAL_QUERY>

=== EARLIER REQUESTS (older first; do not repeat) ===
<EARLIER_REQUESTS>

=== RECENT CONVERSATION (most recent last) ===
<RECENT_CONVERSATION>

=== THIS ROUND'S EVIDENCE ===
<ROUND_EVIDENCE>

=== WORKSPACE BRIEF ===
<WORKSPACE_BRIEF>

=== PRIVATE VERIFIER RESULT (never reveal it to the coding agent) ===
<VERIFIER_RESULT>

=== PRIVATE CONTRACT LEDGER / USER-AGENT STATE ===
<USER_AGENT_STATE>

=== REQUIREMENT-CHANGE POLICY ===
<REQUIREMENT_CHANGE_POLICY>
\end{promptlistingregion}

\begin{promptlistingregion}{promptcardgreenbg}{promptcardgreen}{promptcardgreendark}{OUTPUT CONTRACT}
Return exactly one JSON object:
{
  "decision": "continue | stop | abort",
  "style": "feature_extension | feature_revision | feature_conflict",
  "candidates": [{
    "new_contract": "observable behavior; empty for a revision",
    "supersedes": "replaced contract id for a conflict, else null",
    "preserves": ["active contract ids"],
    "acceptance": {"kind": "observable_output", "oracle": "objective completion condition"},
    "message": "one focused, natural user request"
  }],
  "selected_candidate": 0
}

For a feature revision, candidates has exactly one item that repairs the same contract and adds no new one. For stop or abort, candidates is [] and selected_candidate is null.
\end{promptlistingregion}
\end{structuredprompt}

\section{Verifier Construction Details}
\label{app:verification}

\noindent\textbf{Test rules.} For Single-WS and Cross-WS tasks, the verification agent sees the task specification and the initial workspace, but not the solution. It writes one self-contained pytest file using absolute paths and public interfaces. The tests must exercise the task requirements without network access and without modifying the workspace; temporary files are placed under \path{/tmp}.

\begin{structuredprompt}{Agentic Verifier Construction}
\begin{promptlistingregion}{promptcardbluebg}{promptcardblue}{promptcardbluedark}{ROLE \& INSTRUCTION}
You are a verification engineer agent. A coding task (the "query") has been designed for a future solver, who will work in the repository mounted at /app. The solver has NOT started yet. Your job: produce a hidden, deterministic pytest suite that will decide, after the solver finishes, whether the solver's final workspace satisfies the query's acceptance criteria. The repository at /app is in its INITIAL state. The future solver never sees your tests.
\end{promptlistingregion}

\begin{promptlistingregion}{promptcardbluebg}{promptcardblue}{promptcardbluedark}{HOW TO WORK}
1. Explore the repository (read files; run things where useful) to learn the EXISTING interfaces, file layout, data formats, and packaging.
2. For NEW behavior demanded by the query, take interface details from the query text. Where the query leaves a detail open, assert only what the query actually pins down; never invent stricter requirements than it states.
3. Expected values for the NEW capability must be computed independently inside the test: read input data shipped in the repository at test runtime and recompute the expected result, or take values the query itself states. NEVER derive an expectation for new behavior by running the current, incomplete implementation. For PRESERVATION tests only, the current observed behavior is the baseline.
4. For each task requirement, write 1-3 focused tests. Keep the whole suite at 18 tests or fewer.
5. Do NOT implement preservation by running the repository's own test suite; the solver may legitimately modify those files.
6. Your suite judges the RESULTING BEHAVIOR of the product code, not the solver's process.
\end{promptlistingregion}

\begin{promptlistingregion}{promptcardbluebg}{promptcardblue}{promptcardbluedark}{NON-NEGOTIABLE TEST RULES}
- ONE self-contained file (test_outputs.py): every fixture and helper inside it; no conftest.py or sibling files.
- Absolute paths only (/app/...); NEVER use __file__.
- Public surfaces only: CLI invocations, produced artifacts, documented entry points, config handling.
- Only import the standard library, pytest, and packages already importable (verify with `python3 -c "import X"`); verify executables with `command -v` before use.
- NEVER use pytest.skip/skipif/importorskip/xfail; a missing feature must FAIL.
- Deterministic; no network; no wall-clock or randomness dependence; assertions only.
- Tests may READ /app freely but must never write into it; scratch files go under /tmp.
\end{promptlistingregion}

\begin{promptlistingregion}{promptcardgraybg}{promptcardgray}{promptcardgray}{MANDATORY RED CHECK}
Write your suite to a scratch location and actually run it against the current, unmodified workspace. Requirements on the observed result:
- At least one test that targets the missing capability must FAIL (proving the gap is real and currently open).
- Zero collection errors and zero infrastructure errors; failures must come from assertions or from the expected absence of the new capability, not from a broken test file.
- The preservation tests must PASS now; if one fails on the unmodified workspace, report rejection instead of shipping a broken suite.
Iterate on the suite until these hold, and record the final observed pytest summary.
\end{promptlistingregion}

\begin{promptlistingregion}{promptcardgreenbg}{promptcardgreen}{promptcardgreendark}{OUTPUT CONTRACT}
On success:
{
  "status": "ok",
  "files": {"test_outputs.py": "<the complete, final content of the pytest file>"},
  "initial_run": {"passed": <int>, "failed": <int>, "errors": <int>, "summary": "..."},
  "red_tests": ["names of the tests that fail on the unmodified workspace"],
  "coverage": [{"criterion": "...", "tests": ["test_name", "..."]}],
  "notes": "tolerances, determinism caveats, anything the solve-time harness should know"
}

On rejection:
{
  "status": "rejected",
  "reject_reason": "query_unverifiable | workspace_broken | other",
  "detail": "2-4 sentences of file-level evidence."
}
\end{promptlistingregion}
\end{structuredprompt}

\noindent\textbf{Intent Recovery.} Intent Recovery is not filtered by an agentic verifier, preserving comparability with source-trajectory SFT.

\section{Case Studies}
\label{app:case-studies}

\subsection{Reconstruction and Single-Workspace Re-querying}
\label{app:case-study}

We illustrate reconstruction and reuse with a TypeScript and React project (MirrorBrain).
Its workspace supports several coding tasks without requiring missing source files.
\Cref{tab:case-study-flow} summarizes the reconstruction and re-querying stages.

\begin{table}[H]
\centering
\caption{MirrorBrain reconstruction and re-querying trace.}
\label{tab:case-study-flow}
\begin{tabularx}{\linewidth}{@{}>{\raggedright\arraybackslash}p{0.28\linewidth} >{\raggedright\arraybackslash}X@{}}
\toprule
Stage & Observable artifact \\
\midrule
Source trajectory & Refines React interface components and a focused test across navigation, common controls, review cards, and draft editing while preserving the existing workflows. \\
Deterministic replay & Recovers the project manifest, test configuration, React application, Fastify service, memory-review and cache modules, integration adapters, and end-to-end fixtures. \\
Agentic completion & Preserves all replayed source and test files and adds root-filesystem integration context, including a memory-event fixture, OpenViking adapter, React context, and runtime logs. \\
Intent Recovery & Produces a cross-component interface objective covering navigation, tabs, pagination, candidate cards, and action buttons. \\
Single-WS synthesis & Explores the workspace and generates tasks spanning normalization, URL classification, client errors, API schemas, and cache reconciliation. \\
\bottomrule
\end{tabularx}
\end{table}

\noindent\textbf{Trajectory evidence.}
The source trajectory changes navigation, shared controls, candidate review actions, and
draft views. It also tests keyboard navigation, ARIA state, loading, and disabled behavior.
Replay recovers the surrounding project state; selected paths are shown below.

\begin{treebox}{REPLAYED}{selected paths}{/app/Workspace/mirrorbrain/}
|-- (*@\treejsonnode{package.json}@*)
|-- (*@\treetsnode{vitest.config.ts}@*)
|-- (*@\treefoldernode{tests/e2e/fixtures/}@*)
|   `-- (*@\treetestnode{mirrorbrain-mvp-fixture.ts}@*)
`-- (*@\treefoldernode{src/}@*)
    |-- (*@\treefoldernode{apps/}@*)
    |   |-- (*@\treefoldernode{mirrorbrain-http-server/}@*)
    |   |   |-- (*@\treetsnode{index.ts}@*)
    |   |   `-- (*@\treetestnode{index.test.ts}@*)
    |   `-- (*@\treefoldernode{mirrorbrain-web-react/src/}@*)
    |       |-- (*@\treetsnode{api/client.ts}@*)
    |       |-- (*@\treefoldernode{components/\{artifacts,common,layout,memory,review\}/}@*)
    |       `-- (*@\treetsnode{contexts/MirrorBrainContext.tsx}@*)
    |-- (*@\treefoldernode{integrations/}@*)
    |   |-- (*@\treefoldernode{activitywatch-browser-source/\{index.ts,index.test.ts\}}@*)
    |   `-- (*@\treefoldernode{openviking-store/\{index.ts,index.test.ts\}}@*)
    `-- (*@\treefoldernode{modules/}@*)
        |-- (*@\treefoldernode{memory-events-cache/\{index.ts,index.test.ts\}}@*)
        `-- (*@\treefoldernode{memory-review/\{index.ts,index.test.ts,normalize-memory-items.ts\}}@*)
\end{treebox}

\noindent\textbf{Agentic completion.}
Replay already recovers a functional project. Completion preserves its source and tests
and adds a memory event, integration adapter, React context, and runtime context.
The generated \path{package-lock.json} is omitted.

\begin{treebox}{ADDED}{selected root-filesystem paths}{/}
|-- (*@\treeconfignode{.gitignore}@*)
|-- (*@\treefoldernode{mirrorbrain/memory-events/}@*)
|   `-- (*@\treejsonnode{browser:101.json}@*)
|-- (*@\treefoldernode{src/}@*)
|   |-- (*@\treefoldernode{apps/mirrorbrain-web-react/}@*)
|   |   |-- (*@\treeconfignode{.gitignore}@*)
|   |   `-- (*@\treetsnode{src/contexts/MirrorBrainContext.tsx}@*)
|   `-- (*@\treetsnode{integrations/openviking-store/index.ts}@*)
`-- (*@\treefoldernode{tmp/}@*)
    |-- (*@\treegenericnode{mirrorbrain-debug.log}@*)
    `-- (*@\treegenericnode{mirrorbrain-dev-new.log}@*)
\end{treebox}

\noindent\textbf{Recovered intent.}
Intent Recovery uses the source trajectory, rather than the completed workspace, to recover:

\begin{tcolorbox}[enhanced,breakable,colback=white,colframe=promptrule,
  boxrule=0.35pt,arc=1pt,borderline west={1.2pt}{0pt}{promptaccent},
  left=7pt,right=7pt,top=6pt,bottom=6pt,before skip=6pt,after skip=8pt]
Improve the MirrorBrain React interface by applying a polished, consistent, and
accessible design across main navigation, subtabs, draft tabs, pagination controls,
candidate cards, and action buttons, while preserving existing workflows and component
behavior.
\end{tcolorbox}

The trajectory carries out this objective through consistent active states,
focus-visible styling, keyboard navigation, ARIA semantics, loading states, and a focused
component test.

\noindent\textbf{Generated queries.}
Single-WS ignores the source solution and explores only the completed workspace. Its generated
requests target existing behavior and interfaces:

\begin{enumerate}[leftmargin=1.7em,itemsep=3pt,topsep=3pt]
  \item \textbf{Shell-event normalization.} Extend
  \path{normalize-memory-items.ts} so shell events already captured and rendered by the
  system can participate in daily review without regressing browser events.
  \item \textbf{URL classification.} Replace substring-based development server
  detection with structured URL handling and distinguish GitHub Actions pages from
  ordinary repository pages.
  \item \textbf{Client error handling.} Route five mutation methods in
  \path{api/client.ts} through consistent status-aware JSON handling, including
  descriptive behavior for non-JSON failures.
  \item \textbf{Response schema alignment.} Extend the existing Fastify candidate-memory
  schema with \texttt{role} and \texttt{contribution} so React review components receive
  the required evidence metadata.
  \item \textbf{Cache-source reconciliation.} Merge workspace and OpenViking events at
  cache initialization using existing ID and URL-signature deduplication, preserving
  reverse-chronological order.
\end{enumerate}

\subsection{Cross-WS Synthesis}
\label{app:cross-workspace-case-study}

We trace Cross-WS synthesis on two C toolkits for RSA cryptography. The target provides key generation,
encryption, and signing but no decryption program; the reference implements the full
round trip. Relation matching therefore selects RSA decryption as the missing capability.
\Cref{tab:cross-workspace-case-flow} summarizes the synthesis stages.

\begin{table}[H]
\centering
\caption{Synthesis trace for the RSA toolkit dependency pair.}
\label{tab:cross-workspace-case-flow}
\begin{tabularx}{\linewidth}{@{}>{\raggedright\arraybackslash}p{0.27\linewidth} >{\raggedright\arraybackslash}X@{}}
\toprule
Stage and output & Evidence \\
\midrule
Relation mining & Independently generated profiles
classify both workspaces as C RSA toolkits; TF--IDF retrieval proposes the pair, and
the relation judge records the directional capability gap ``RSA decryption''. \\
Pair packaging & The target is mounted writable at
\texttt{/app/workspaces/repo1}; the reference is mounted read-only at
\texttt{/app/workspaces/repo0}. \\
Query generation & The generator reads the actual
files of both workspaces, confirms that the capability is implemented in the reference
and absent from the target, and generates a task grounded in facts found only in the
reference. \\
Verifier generation & An in-container agent authors the test suite and executes it on
the initial workspace. New-capability tests fail while the preservation tests and the
reference-unchanged check pass, satisfying the red-check requirement. \\
Solution rollout & The teacher solves the task; all
clean checks pass, and the trajectory is retained in the pass-only corpus. \\
\bottomrule
\end{tabularx}
\end{table}

\noindent\textbf{Mounted pair.}
The two toolkits share a domain but differ in completeness:

\begin{treebox}{SOLVER}{read-only reference and writable target \textperiodcentered\ key files}
  {/app/workspaces/}
|-- (*@\treefoldernode{repo0/}@*)
|   |-- (*@\treetsnode{rsa\_keygen.c}@*)
|   |-- (*@\treetsnode{rsa\_encrypt.c}@*)
|   |-- (*@\treetsnode{rsa\_decrypt.c}@*)
|   |-- (*@\treetsnode{rsa\_sign.c}@*)
|   |-- (*@\treefoldernode{input/}@*)
|   |   `-- (*@\treegenericnode{plaintext.txt}@*)
|   `-- (*@\treefoldernode{output/}@*)
|       `-- (*@\treegenericnode{decrypted.txt}@*)
`-- (*@\treefoldernode{repo1/}@*)
    |-- (*@\treetsnode{rsa\_keygen.c}@*)
    |-- (*@\treetsnode{rsa\_encrypt.c}@*)
    |-- (*@\treetsnode{rsa\_sign.c}@*)
    |-- (*@\treefoldernode{input/}@*)
    |   `-- (*@\treegenericnode{plaintext.txt}@*)
    |-- (*@\treefoldernode{output/}@*)
    |   `-- (*@\treegenericnode{decrypted.txt}@*)
    `-- (*@\treefoldernode{task\_file/}@*)
        |-- (*@\treefoldernode{input/}@*)
        |   `-- (*@\treegenericnode{plaintext.txt}@*)
        `-- (*@\treefoldernode{output/}@*)
\end{treebox}

\noindent\textbf{Grounded capability gap.}
The target has key generation, encryption, and signing but no decryption source. Its
encryptor fixes the padding mode and public-key loader, while the plaintext is
the round-trip input. The reference reveals the private-key loader, decrypt
call, padding mode, and hardcoded input paths. These hidden facts do not appear in the
task, which names only the reference root.

\noindent\textbf{Selected request (abridged).}
\begin{tcolorbox}[enhanced,breakable,colback=white,colframe=promptrule,
  boxrule=0.35pt,arc=1pt,borderline west={1.2pt}{0pt}{promptaccent},
  left=7pt,right=7pt,top=6pt,bottom=6pt,before skip=6pt,after skip=8pt]
The RSA cryptography toolkit at \texttt{/app/workspaces/repo1} currently supports key
generation, encryption, and signing, but is missing the decryption program needed to
complete the encrypt/decrypt round-trip. Create a new file \texttt{rsa\_decrypt.c} in
the root of \texttt{/app/workspaces/repo1} that reads an encrypted binary file and a
private key, then decrypts the ciphertext back to plaintext. The program must
integrate seamlessly with the existing toolkit: it should work with keys produced by
the existing \texttt{rsa\_keygen.c} and ciphertext produced by the existing
\texttt{rsa\_encrypt.c}, and it must follow the coding conventions, error-handling
style, OpenSSL API usage patterns, and output path conventions already established by
the other programs. A reference implementation exists at
\texttt{/app/workspaces/repo0}: consult it to determine the exact OpenSSL API calls,
padding mode, PEM reading function, file paths, command-line interface, and resource
cleanup pattern required. When run after \texttt{rsa\_keygen} and
\texttt{rsa\_encrypt}, the program must produce an output file whose contents are
byte-for-byte identical to the original plaintext input.
\end{tcolorbox}

\noindent\textbf{What the verifier checks.}
The suite covers the new file, compilation, round trips, command-line behavior, input
paths, padding, and OpenSSL calls. Preservation tests cover the existing programs and
keep the reference unchanged. The teacher solution passes the full suite.

\subsection{Multi-Round User Queries}
\label{app:user-continuation-case-study}

\begin{figure}[H]
\centering
\includegraphics[width=\textwidth]{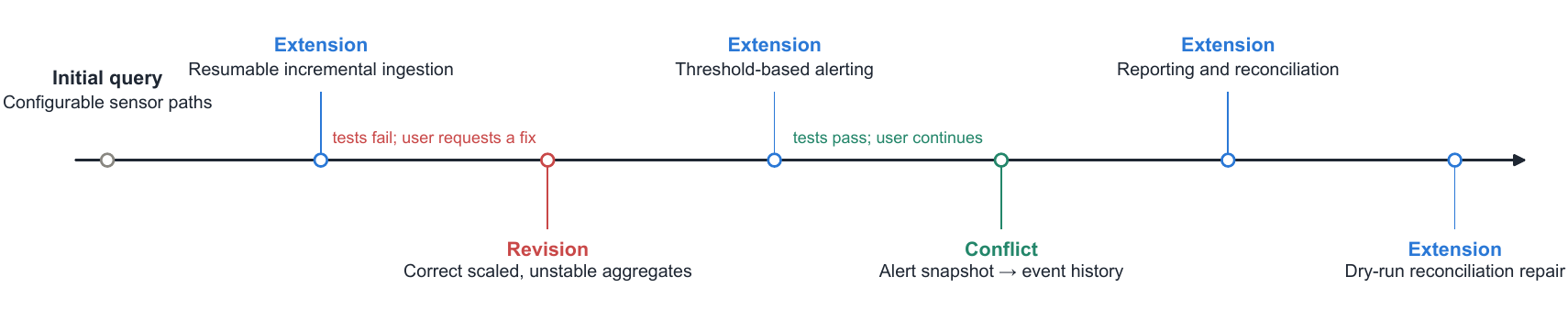}
\caption{\textbf{Verifier-guided multi-round continuation.} A private verifier checks
each round, while the coding agent sees only the subsequent natural-language
requests.}
\label{fig:q4-trajectory}
\end{figure}

\begin{table}[H]
\centering
\small
\caption{Distribution of interaction styles across continuation rounds in the retained Multi-Round records. We impose no target ratio over the styles.}
\label{tab:q4-style-distribution}
\setlength{\tabcolsep}{6pt}
\begin{tabularx}{0.60\textwidth}{@{}>{\raggedright\arraybackslash}X c@{}}
\toprule
Interaction style & Share \\
\midrule
Feature extension & 62.7\% \\
Feature revision & 29.3\% \\
Feature conflict & 8.0\% \\
\bottomrule
\end{tabularx}
\end{table}

\Cref{tab:user-continuation-flow} follows the user requests across rounds.

\begin{table}[H]
\centering
\caption{Abridged user requests from the Multi-Round sensor-ingestion trajectory. Bracketed
ellipses mark omitted text. ``Revision''
corrects the current requirement; ``conflict'' replaces it.}
\label{tab:user-continuation-flow}
\begin{tabularx}{\linewidth}{@{}>{\raggedright\arraybackslash}p{0.20\linewidth} >{\raggedright\arraybackslash}X@{}}
\toprule
Round & User-request excerpt \\
\midrule
Initial request & ``Working in the project root \texttt{/app}, the entrypoint
\texttt{sensor\_processor.py} hard-codes both its input path ... and its output path ...,
which makes the script impossible to run against alternate data without editing source.
Make it possible to invoke the script as \texttt{python sensor\_processor.py <input\_csv>
<output\_json>} so that the two positional arguments override the input and output
locations.'' \\
1: Extension & ``The path override work is solid, thanks. Now I need this to stop being a
one-shot script, because in production my \texttt{sensor\_data.csv} keeps growing and I
don't want to re-crunch the entire file every time I get a new batch of readings. [...]
Please add a resumable/incremental ingestion mode backed by a small persisted state store.'' \\
2: Revision & ``The row-counting side of the incremental mode looks right ---
\texttt{readings\_count} comes out at 150 per sensor on a fresh run and rises to exactly
154 after I append four rows, so old rows aren't being double-counted. But the actual
aggregate values are wrong and unstable, and it's making the output useless for me. [...]
It looks like each reading is being scaled down before it's aggregated.'' \\
3: Extension & ``The aggregates are solid and stable now --- thanks. The next thing I actually
need for production is alerting, because right now I have to eyeball
\texttt{processed\_results.json} to notice when a sensor drifts out of its safe range. [...]
Please add a threshold rules engine in a new \texttt{rules.py}, driven by a JSON rules config.'' \\
4: Conflict & ``The alerting engine is doing its job now, but I've hit a real production gap:
I've decided to change how \texttt{alerts.json} works. Right now it's overwritten every
run holding only the new alerts for that run, which means when a sensor recovers the alert
just silently vanishes and I have no record it ever fired or when it cleared. [...]
Please turn \texttt{alerts.json} into an append-only history of transition events.'' \\
5: Extension & ``The event history is finally durable, but right now \texttt{alerts.json} and
\texttt{alerts.jsonl} are write-only --- nothing ever reads them back, so when I come in
Monday morning I still have to eyeball a growing history file to answer basic questions.
I need a reporting command that turns that trail into an operational summary.'' \\
6: Extension & ``The reconcile check has already caught me out once --- it correctly flagged
that \texttt{state.json} and the \texttt{alerts.json} history disagreed about what was open
--- but all it can do is tell me there's drift and exit non-zero. [...] I need reconcile
to be able to actually fix the drift, not just report it. Please add a repair capability
on top of the existing reconcile logic.'' \\
\bottomrule
\end{tabularx}
\end{table}

\noindent\textbf{Failure-driven feature revision.}
The first continuation adds persistent incremental aggregation, but unseeded random
scaling makes its values unstable. The private verifier catches this failure. The user agent keeps
the same requirements and asks for a correction without exposing the tests:

\begin{tcolorbox}[enhanced,breakable,colback=white,colframe=promptrule,
  boxrule=0.35pt,arc=1pt,borderline west={1.2pt}{0pt}{q4revisionborder},
  left=7pt,right=7pt,top=6pt,bottom=6pt,before skip=6pt,after skip=8pt]
The row counts are right, but the aggregate values are wrong and unstable. The averages,
minima, and maxima are about half of the readings in the CSV, and rerunning unchanged
input changes the result. Please make the aggregates use the real readings and remain
deterministic for a fresh run, reset, and no-op rerun, while preserving the record shape,
rounding, ordering, and incremental counts.
\end{tcolorbox}

The agent removes the random scaling and passes the next check, after which the user agent requests
threshold-based alerting. Test code and results never enter the solver-visible conversation.

\noindent\textbf{Controlled requirement conflict.}
After alerting passes, the user agent replaces the per-run JSON list with a durable event
history, stating this as a new product decision rather than a defect:

\begin{tcolorbox}[enhanced,breakable,colback=white,colframe=promptrule,
  boxrule=0.35pt,arc=1pt,borderline west={1.2pt}{0pt}{q4conflictborder},
  left=7pt,right=7pt,top=6pt,bottom=6pt,before skip=6pt,after skip=8pt]
I have decided to change how \texttt{alerts.json} works. It should no longer be a
per-run snapshot: make it an append-only history of firing and resolved transitions, so
a recovered alert remains recorded. Keep exit code 2 for a new firing and 0 for a lone
resolution, and preserve the first-observed timestamp and severity with every event.
\end{tcolorbox}

The revised history requirement governs the remaining rounds, which add reporting,
reconciliation, and repair.

\section{SFT Dataset Statistics}
\label{app:sftstats}

\Cref{tab:sftstats} summarizes the final terminal SFT corpora, with turns, tool calls, and tokens reported as per-record medians. Multi-Round records are substantially longer because they extend a completed session with follow-up user requests.

\begin{table}[H]
\centering
\caption{Final SFT corpus statistics after length filtering.}
\label{tab:sftstats}
{\small
\setlength{\tabcolsep}{2pt}
\renewcommand{\arraystretch}{1.18}
\begin{tabularx}{0.72\textwidth}{@{}>{\raggedright\arraybackslash}X c c c c@{}}
\toprule
\textbf{Dataset} & \textbf{Records} & \textbf{Turns} & \textbf{Tool calls} & \textbf{Tokens} \\
\midrule
Intent Recovery & 35{,}809 & 12 & 17 & 36.1k \\
\rowcolor{tblzebra} Single-WS & 25{,}386 & 14 & 20 & 30.4k \\
Cross-WS & 3{,}512 & 23 & 38 & 46.5k \\
\rowcolor{tblzebra} Multi-Round & 3{,}079 & 92 & 102 & 126.1k \\
\bottomrule
\end{tabularx}
}
\end{table}

\section{Evaluation Configurations}
\label{app:openweights-eval-config}

\Cref{tab:openweights-eval-config} lists the configurations used for Terminal-Universe-27B and the five reproduced open-weight baselines in \Cref{tab:main}. These reproduced baselines use model-specific settings rather than a shared evaluation configuration. The listed settings apply to both Terminal-Bench 2.1 and EvoCode-Bench v2 unless noted otherwise. For TMax-27B, the listed configuration applies only to our EvoCode-Bench v2 reproduction; its Terminal-Bench scores and Vanillux2Agent scaffold follow the original report. Qwen3.5-27B and Qwen3.7-Max use the Terminal-Universe-27B configuration, with xhigh effort enabled for Qwen3.7-Max. On EvoCode-Bench v2, TermiGen-32B instead uses Terminus2-JSON with an 81{,}920-token summarization threshold.

\begin{table}[H]
\centering
\caption{Evaluation configurations used in our reproductions.}
\label{tab:openweights-eval-config}
{\scriptsize
\setlength{\tabcolsep}{2pt}
\renewcommand{\arraystretch}{1.14}
\begin{tabularx}{\textwidth}{@{}>{\raggedright\arraybackslash}p{2.5cm} *{6}{>{\centering\arraybackslash}X}@{}}
\toprule
\textbf{Setting} & \makecell{\textbf{Nemotron-}\\\textbf{Terminal-32B}} & \makecell{\textbf{OpenThinker-}\\\textbf{Agent-32B}} & \textbf{TermiGen-32B} & \makecell{\textbf{TerminalTraj-}\\\textbf{32B}} & \textbf{TMax-27B} & \makecell{\textbf{Terminal-}\\\textbf{Universe-27B}} \\
\midrule
Architecture & \makecell{\texttt{Qwen3For}\\\texttt{CausalLM}} & \makecell{\texttt{Qwen3For}\\\texttt{CausalLM}} & \makecell{\texttt{Qwen2For}\\\texttt{CausalLM}} & \makecell{\texttt{Qwen2For}\\\texttt{CausalLM}} & \makecell{\texttt{Qwen3\_5For}\\\texttt{Conditional}\\\texttt{Generation}} & \makecell{\texttt{Qwen3\_5For}\\\texttt{Conditional}\\\texttt{Generation}} \\
\rowcolor{tblzebra} Scaffold & Terminus2 & Terminus2 & BashAgent & Terminus2 & Terminus2 & Terminus2 \\
Parser & JSON & JSON & Text JSON & JSON & XML & XML \\
\rowcolor{tblzebra} Context length & 40{,}960 & 32{,}768 & 131{,}072 & 32{,}768 & 262{,}144 & 262{,}144 \\
Turn max tokens & 8{,}192 & 8{,}192 & 8{,}192 & 8{,}192 & 65{,}536 & 65{,}536 \\
\rowcolor{tblzebra} Summ. threshold & 32{,}960 & 28{,}672 & disabled & 24{,}768 & 176k & 176k \\
Temperature & 0.6 & 0.6 & 0.6 & 0.7 & 0.7 & 1.0 \\
\rowcolor{tblzebra} Top-$p$ / top-$k$ & 0.95 / 20 & 0.95 / 20 & 0.8 / 20 & 0.8 / 20 & 0.95 / 20 & 0.95 / 20 \\
Interleaved thinking & \tno & \tno & \tno & \tno & \tyes & \tyes \\
\rowcolor{tblzebra} Thinking enabled & \tyes & \tyes & \tno & \tno & \tyes & \tyes \\
\midrule
\rowcolor{tblbandgray} \tblock{7}{promptcardgray}{Shared across all reproductions} \\
TB wall clock & \multicolumn{6}{c}{4 h} \\
\rowcolor{tblzebra} EvoCode wall clock & \multicolumn{6}{c}{10 h} \\
TB runs & \multicolumn{6}{c}{6} \\
\rowcolor{tblzebra} EvoCode runs & \multicolumn{6}{c}{4} \\
\bottomrule
\end{tabularx}
}
\end{table}

\end{document}